\documentclass[5p]{elsarticle}
\usepackage{amsmath}
\usepackage{amssymb}
\usepackage{graphicx}
\usepackage{booktabs}
\usepackage{algorithmic}
\usepackage{bm}
\usepackage{subfigure} 
\usepackage{float}
\usepackage{mwe}
\usepackage{color}
\usepackage[colorlinks=true, linkcolor=blue, anchorcolor=blue, citecolor=blue]{hyperref}
\usepackage{multirow}

\usepackage{url}
\title{A Comprehensive Approach for UAV Small Object Detection with Simulation-based Transfer Learning and Adaptive Fusion\tnoteref{t1,t2}}

\author{Chen Rui, Guo Youwei, Zheng Huafei, Jiang Hongyu\corref{mycorrespondingauthor}}
\address{Institute of Electronic Engineering, China Academy of Engineering Physics, MianYang, 621999, China}
\cortext[mycorrespondingauthor]{Corresponding author: doherty2004@163.com}

\begin{document}
	
	\begin{abstract}
		Precisely detection of Unmanned Aerial Vehicles(UAVs) plays a critical role in UAV defense systems. Deep learning is widely adopted for UAV object detection whereas researches on this topic are limited by the amount of dataset and small scale of UAV. To tackle these problems, a novel comprehensive approach that combines transfer learning based on simulation data and adaptive fusion is proposed. Firstly, the open-source plugin AirSim proposed by Microsoft is used to generate mass realistic simulation data. Secondly, transfer learning is applied to obtain a pre-trained YOLOv5 model on the simulated dataset and fine-tuned model on the real-world dataset. Finally, an adaptive fusion mechanism is proposed to further improve small object detection performance. Experiment results demonstrate the effectiveness of simulation-based transfer learning which leads to a 2.7\% performance increase on UAV object detection. Furthermore, with transfer learning and adaptive fusion mechanism, 7.1\% improvement is achieved compared to the original YOLO v5 model.
	\end{abstract}
	
	
	\maketitle
	
	\section{Introduction}
	In recent years, Unmanned Aerial Vehicles (UAVs) have been widely employed in agriculture, industry, military, and other fields, bringing not only great convenience to producing and living but also great security risks such as trespass, illegal delivery, spy flying, etc. Therefore, researches on the detection and recognition of UAV small targets has been rapidly growing. Currently, UAV object detection technology is commonly based on deep learning object detection algorithms  \cite{zhang2018spatial}\cite{tong2019uav}\cite{9203924}. Although deep learning methods have made great progress in general object detection, some popular detectors such as Fast R-CNN \cite{girshick2015fast}, Faster R-CNN \cite{ren2015faster}, SSD \cite{liu2016ssd}, YOLO \cite{redmon2016you}, RetinaNet \cite{lin2017focal} still work poorly in small object detection tasks including UAV detection \cite{guvenc2018detection}, pedestrian detection \cite{brunetti2018computer}, traffic sign detection \cite{2012Vision}, etc. For instance, accuracy of the state-of-the-art model on small objects($AP_s$) is only 43.9\% \cite{dai2021dynamic}, where the "small object" is defined referring to MS COCO dataset \cite{lin2014microsoft}. 
	
	We reveal two main reasons for the poor performance of deep learning based detectors. One reason is the limited quantity of supervised data. 
	Deep learning algorithms are data-consuming and will converge well only if sufficient training samples are provided. Unfortunately supervised data are always insufficient. Taking UAV as targets for example, it's expensive to collect enough UAV flying images which requires licensed pilots, a variety of drones, and high-def cameras. Consequently, massive manual work is necessary to annotate images collected. Another reason is the small pixel size of UAV object in images. Features of small objects gradually disperse or even vanish as passed through layers of neural network such as ConvNet, which always leads to bad results output from detectors. Data augmentation and multi-scale feature learning are two methods widely used in small object detectors to solve these problems respectively. Kisantal et al. \cite{kisantal2019augmentation} proposed a data augmentation method by oversample images with small objects and copy-pasting small objects many times. Li et al. \cite{li2017fssd} presented feature fusion single shot multi-box detector (FSSD), which adopted a new lightweight feature fusion module which improved the small object detection performance over SSD. Deng et al. \cite{deng2021extended} proposed the extended feature pyramid network (EFPN), which introduced an extra high-resolution pyramid level specialized for small object detection. Based on the above analysis, we try to improve UAV small object detection from both aspects of data augmentation and multi-scale feature learning.
	
	Firstly, to deal with the limited quantity of supervised data, we collect real-world images containing UAV objects from Internet and spend considerable human resources on annotation. Furthermore, we construct a simulated UAV dataset covering 12 scenes and four common UAV models as the complement of the real-world images. However, there is a gap between reality and simulation which hinders the performance of detectors and cannot be ignored if we utilize both the real-world and the simulated dataset. Therefore, we secondly introduce transfer learning to minish the gap. Namely, we take the simulated data as the auxiliary training samples on which a YOLO v5 model is pretrained expecting the general knowledge for UAV small object detection would be obtained. After that the pretrained model will be fine-tuned on real-world dataset which is regarded as the target domain. Compared with the model trained solely on real-world dataset, our model trained by transfer learning achieves an increase of 2.7 percentage points in $AP$. Thirdly, to make our model more suitable for small object detection, we introduce adaptive fusion mechanism aiming at a better fusion of shallow and deep features into the neck block of YOLO v5. Experimental results show that compared with the original YOLO v5, our model gains a 7.1\% improvement in $AP$, which means appropriately suppressing the propogation of deep features is better for small object detection.
	
\begin{figure*}[htb]
		\centering
		\includegraphics[width=2\columnwidth]{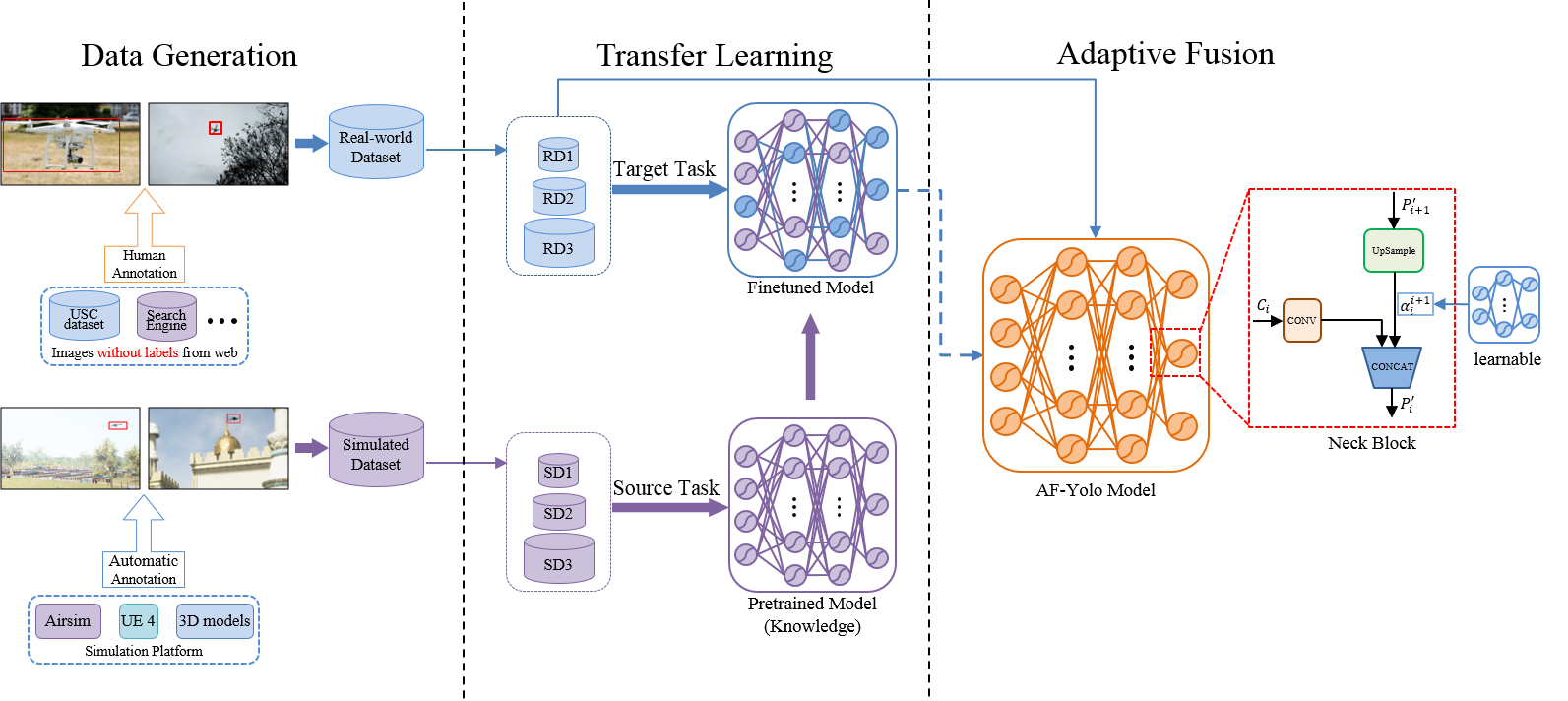}
		\centering
		\caption{Illustration of the approach proposed in this paper.}
		\label{bigchart}
	\end{figure*}

	The contributions of our work is illustrated by Figure \ref{bigchart} and can be summarized as following:\\
	\begin{itemize}
		\item
		\sloppy
		We construct and publicly release two kinds of datasets: the real-world dataset(RD) and the simulated dataset(SD). The RD is annotated manually which consists of images collected from Internet. The SD is generated with high fidelity by simulation platform aiming to alleviate the demand of supervised data.
		
		\item
		We introduce transfer learning to improve performance on real-world dataset (target domain) by transferring general knowledge learned from simulated dataset (source domain) even in case of low-resource for real-world dataset.
		
		\item
		We propose a model called AF-YOLO by bringing adaptive fusion into the neck part of YOLO v5. The modified model achieves a better fusion of shallow features and deep features.
	\end{itemize}
	
	The rest of the paper is organized as follows. Section \ref{relatedwork} provides an overview of related work. Section \ref{methods} describes our methods of data generation, transfer learning, and modification of YOLO v5. Elaborate experiments with analysis are presented in section \ref{experiment}. In detail, subsection \ref{exp_dataset} proves the validation of simulated dataset, subsection \ref{exp_tl} demonstrates the improvement obtained by transfer learning even on a small part of real-world dataset, and subsection \ref{exp_af} renders a better performance by AF-YOLO. The paper is conclued with future work in Section \ref{conclusion}.
	
	\section{Related Work}
	\label{relatedwork}
	\subsection{Dataset for UAV Small Object Detection}
	To deal with the detection of common objects, many datasets such as MS COCO, PASCAL VOC \cite{everingham2015pascal}, and ImageNet \cite{deng2009imagenet} have been released. However, few of existing datasets are special for UAV object detection, especially small objects. TrackingNet \cite{muller2018trackingnet} is an general object tracking dataset that consists of 30511 video streamings picked from YouTube-BB dataset\cite{real2017youtube}. However UAV objects only account for 5\% in TrackingNet. LaSOT\cite{fan2019lasot}, another tracking video dataset for general categories, collects 1400 videos manually, but only 20 video streams containing drone targets which are mostly in medium or large size. USC dataset \cite{chen2017deep} is a drone detection and tracking dataset containing 30 video clips shot at the USC campus. Although all videos of USC dataset are shot with a wide range of background scenes, only a single drone model were used. Besides, there is a common problem of above three datasets that the size distribution of UAV targets is not uniform.
	
	In our work, we utilize two ideas to cope with insufficiency of dataset. First, we collect images existing in the Internet or other video tracking datasets(e.g. USC dataset) and annotate them if not labeled. Second, instead of copy-pasting, we apply a simulation platform to generate more realistic images which covering various scenes and UAV models and allow several UAVs appear in one image.

	\subsection{Transfer Learning}
	In most tasks of machine learning, we assume that the data used during training(assuming the data distribution is $P$) and inference(assuming the distribution is $Q$) obey a similar distribution(meaning $P=Q$). Nevertheless, this assumption is difficult to meet in practice, meaning the data has different distributions($P\neq Q$). The research of transfer learning try to alleviated this dilemma. Generally, data in source domain is easy to obtain. In contrast, data in target domain is scarce. So transfer learning studies how to exploit abundant data in source domain along with sparse data in target domain to train a model suitable for target domain \cite{9134370}. Transfer learning has already applied in fields of activity recognition \cite{cook2013transfer}, visual categorization \cite{6847217}, deep visual domain adaption \cite{WANG2018135}, and sentiment analysis \cite{8746210}. As for object detection, Zhong et al. \cite{chen2018end} use the transfer learning method to train a deep convolutional neural network on limited samples for airplane detection. Talukdar et al. \cite{8474198} pre-train a convolutional neural networks on synthetic images and obtain a promising improvement of object detection. M.Nalamati et al. \cite{8909830} and D. T. Wei Xun et al. \cite{9358449} also use transfer learning to reduce the need for real-world data.
	
	In our method, given real-world dataset and simulated dataset, we naturally introduce transfer learning to train a detector which is suitable for real-world UAV small objects eventually.
	
	\subsection{Multi-scale Feature Fusion}
	Feature maps at different stages represent different levels of abstract information. Shallow feature maps have small receptive fields but rich information about location and edges, which is beneficial for small target detection; deep feature maps have large receptive fields and rich semantic information, which is appropriate for detecting larger targets. Therefore, feature maps at different stages can be fused to improve the performance of multi-scale target detection. Since FPN(Feature Pyramid Networks) \cite{lin2017feature} was proposed in 2017, the process of feature fusion as a necessity has been added to various detectors, and different algorithms have been merged for feature fusion. Path Aggregation Network(PANet) \cite{liu2018path} introduces a bottom-up path augmentation based on FPN. Multi-level Feature Pyramid Network(MLFPN) \cite{zhao2019m2det} firstly extracts more representative multi-level and multi-scale features by means of thinned u-shape modules and feature fusion modules, and secondly fuses multi-level features using scale feature aggregation module. Bi-FPN \cite{Tan_2020_CVPR} implements both top-down and bottom-up feature fusion and introduces weight in the feature fusion stage.
	
	In this paper, for the specific task of UAV small objection detection, we apply multi-scale feature fusion by means of adaptive fusion to properly suppress the influence of deep features which concern large targets more.
	
	\begin{figure*}[htb]
		\centering
		\includegraphics[width=2\columnwidth]{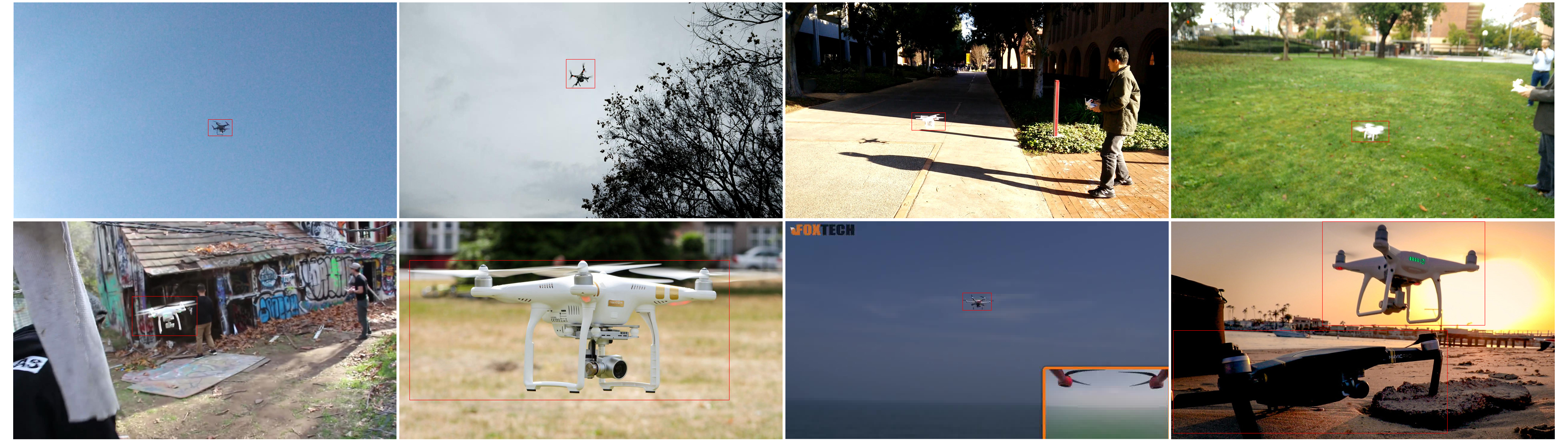}
		\centering
		\caption{Examples of the real-world dataset.}
		\label{examples_RD}
	\end{figure*}
	
	\section{Methods}
	\label{methods}
	Based on general object detection tasks, here we give a formal task description of UAV small object detection in this paper: to determine whether there are any instances of UAV,  specifically multi-rotors, in a image and, if present, to return the spatial location and extent of each UAV instance via a bounding box. The term of "small" means the ground-truth box size of each UAV instance is always small, exactly no larger than 32*32 pixels in a 640*640 image.
	
	The overall architecture of our methods is illustrated in \ref{bigchart}, which consists of process of data generation, transfer learning approach, and modification of YOLO v5. Detail descriptions of our methods are presented as following.

	\subsection{Data Generation}
	For most data-driven deep learning algorithms, resource of dataset plays a significant role. As mentioned above, it is often difficult to obtain enough UAV small object datasets. To address this issue, we construct two kinds of datasets: real-world dataset(RD) and simulated dataset(SD). The real-world dataset consists of images collected from other video streaming dataset or Internet. The simulation dataset, called SimUAV in this paper, is generated by simulation platform to alleviate the data-insufficiency further. Both datasets are released on github \footnote{Dataset webpage: https://github.com/mlcaepiee/SimUAV}, where detailed information could be found.
	\subsubsection{Real-world Dataset}
	In reality, it is often difficult to obtain enough UAV small target datasets with diverse backgrounds and UAV models. In this paper, we construct a real-world dataset containing 21803 images, including 17853 images from USC dataset \cite{chen2017deep} and others from Amateur Unmanned Air Vehicle Detection Dataset \cite{web:AmateurUAV} (2863 images)and Kaggel Drone Dataset \cite{web:MehdiUAV}(1087 images). Figure \ref{examples_RD} shows several examples in real-world dataset. 
	
\begin{figure}[H]
		\centering
		\includegraphics[width=1\columnwidth]{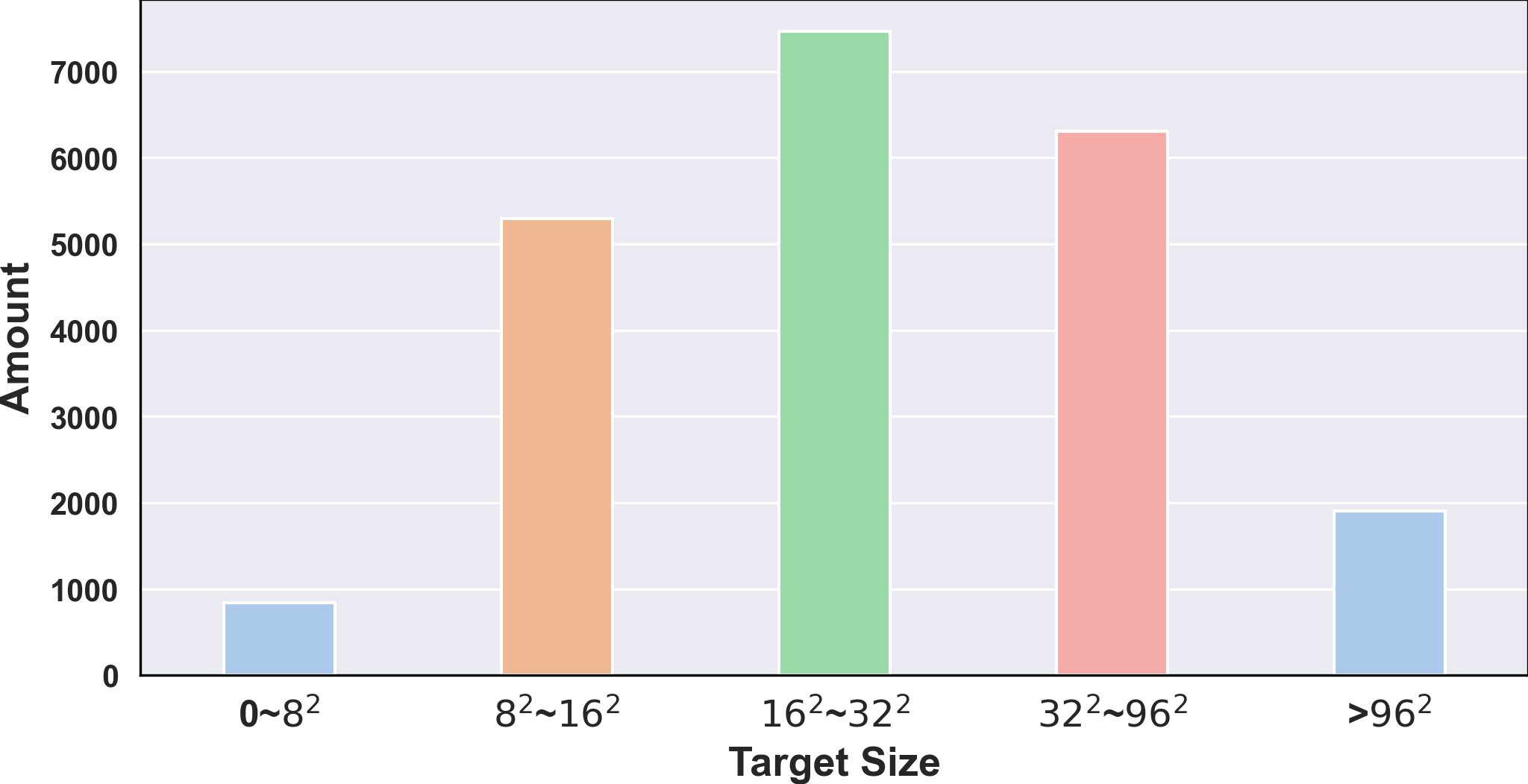}
		\centering
		\caption{Target size distribution in real-world dataset.}
		\label{dist_RD}
	\end{figure}

	Since images are collected and searched without limitation on target size, there is no guarantee that targets of real-world dataset obey the "small-size" constraint inevitably. Figure \ref{dist_RD} shows the uneven distribution of UAV target sizes. Note that small targets(size$\leq32^2$) take 53.5\% and even large targets(size$\geq96^2$) occupy a certain proportion of the dataset. Considering UAV small object detection as our task, we construct a test dataset containing only small targets from real-world dataset and inference our models on this dataset. Meanwhile, we remain large targets in train dataset to prevent the problem of generalization.
	
	\subsubsection{Simulation Dataset}
\begin{figure}[htbp]
		\centering
		\includegraphics[width=1\columnwidth]{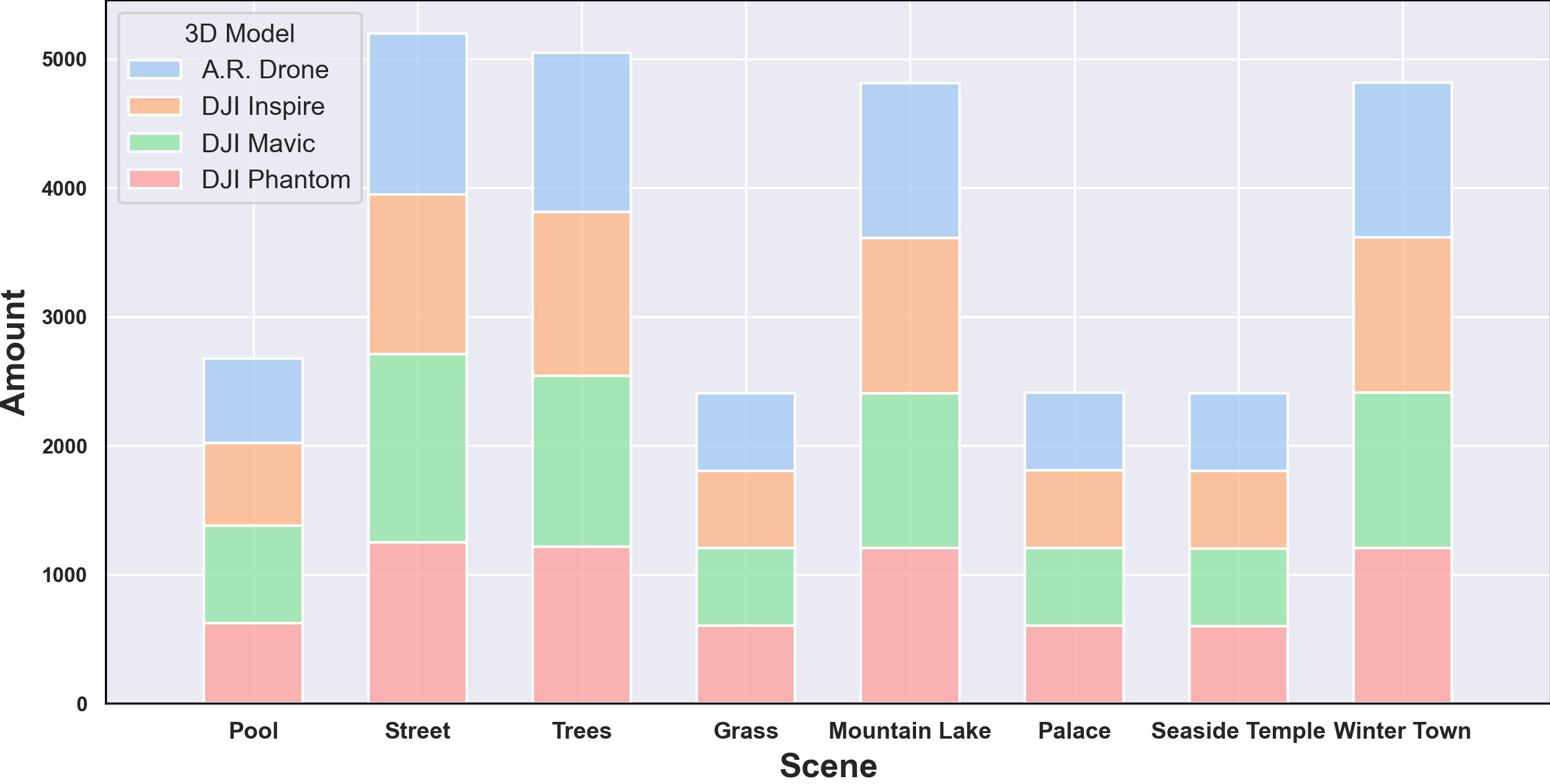}
		\centering
		\caption{Distribution of scenes and models in simulated dataset.}
		\label{sim_prop}
	\end{figure}

\begin{figure*}[t]
		\centering
		\subfigure[Pool]{
			\centering
			\includegraphics[width=0.35\columnwidth, height=0.3\columnwidth]{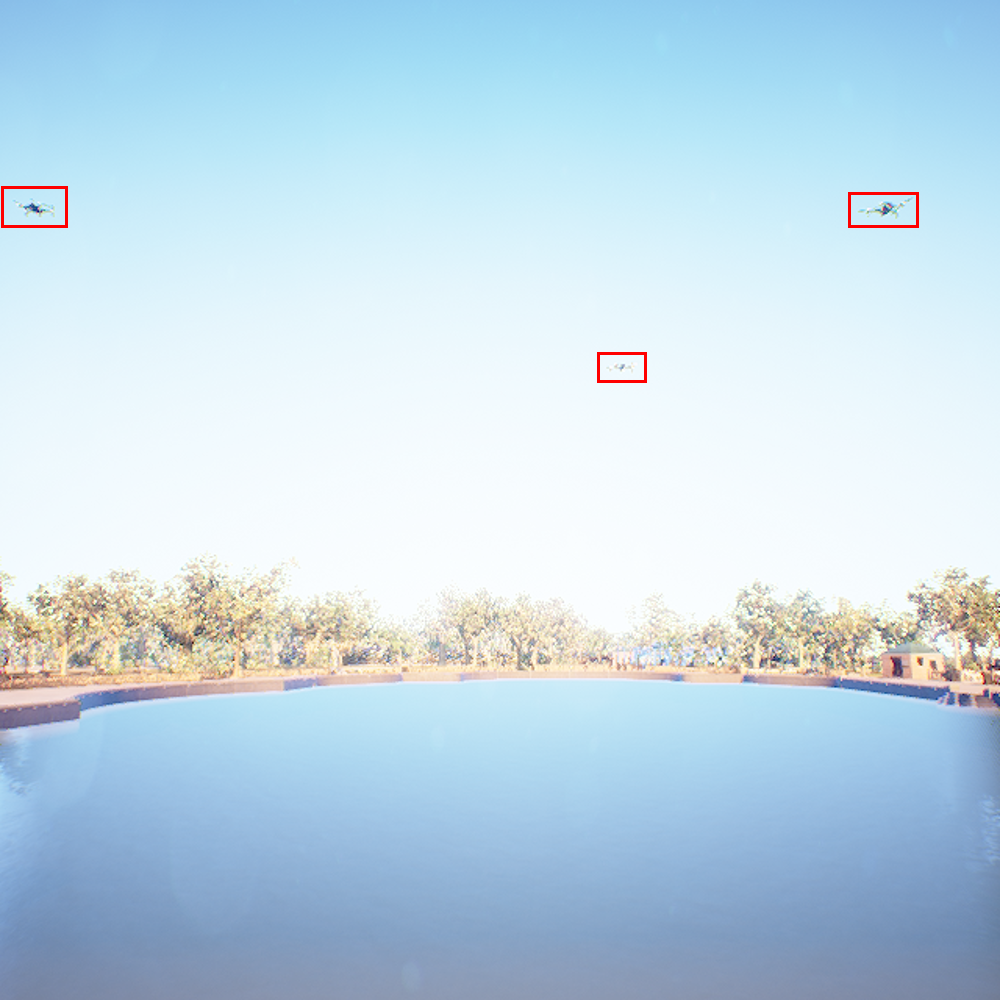}
		}
		\subfigure[Street]{
			\centering
			\includegraphics[width=0.35\columnwidth, height=0.3\columnwidth]{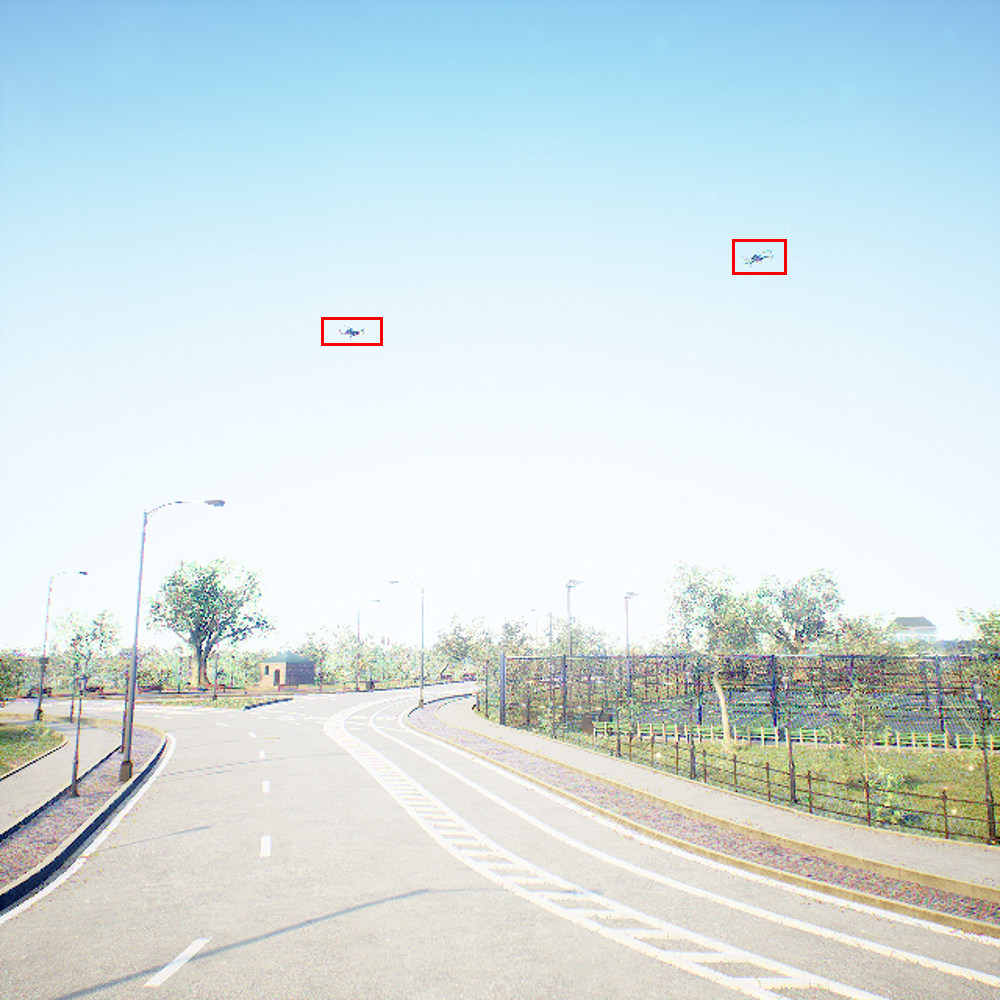}
		}
		\subfigure[Trees]{
			\centering
			\includegraphics[width=0.35\columnwidth, height=0.3\columnwidth]{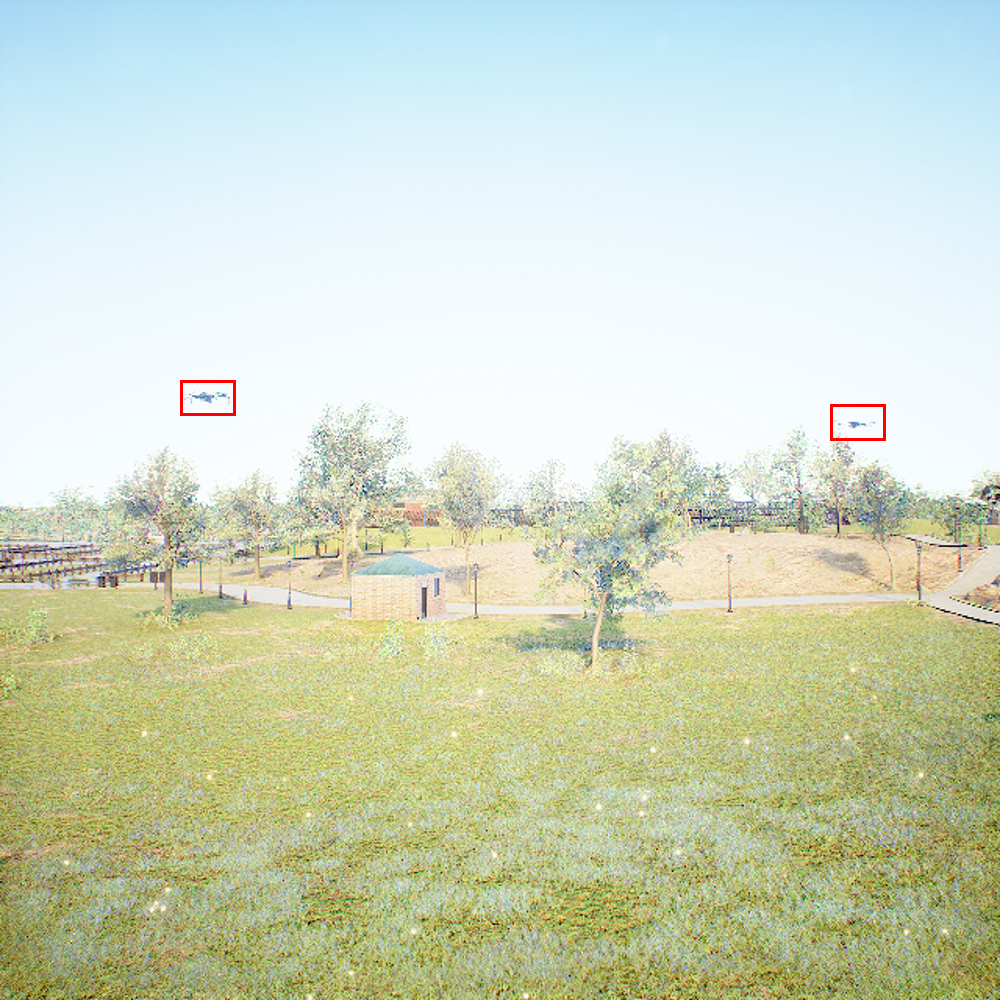}
		}
		\subfigure[Grass]{
			\centering
			\includegraphics[width=0.35\columnwidth, height=0.3\columnwidth]{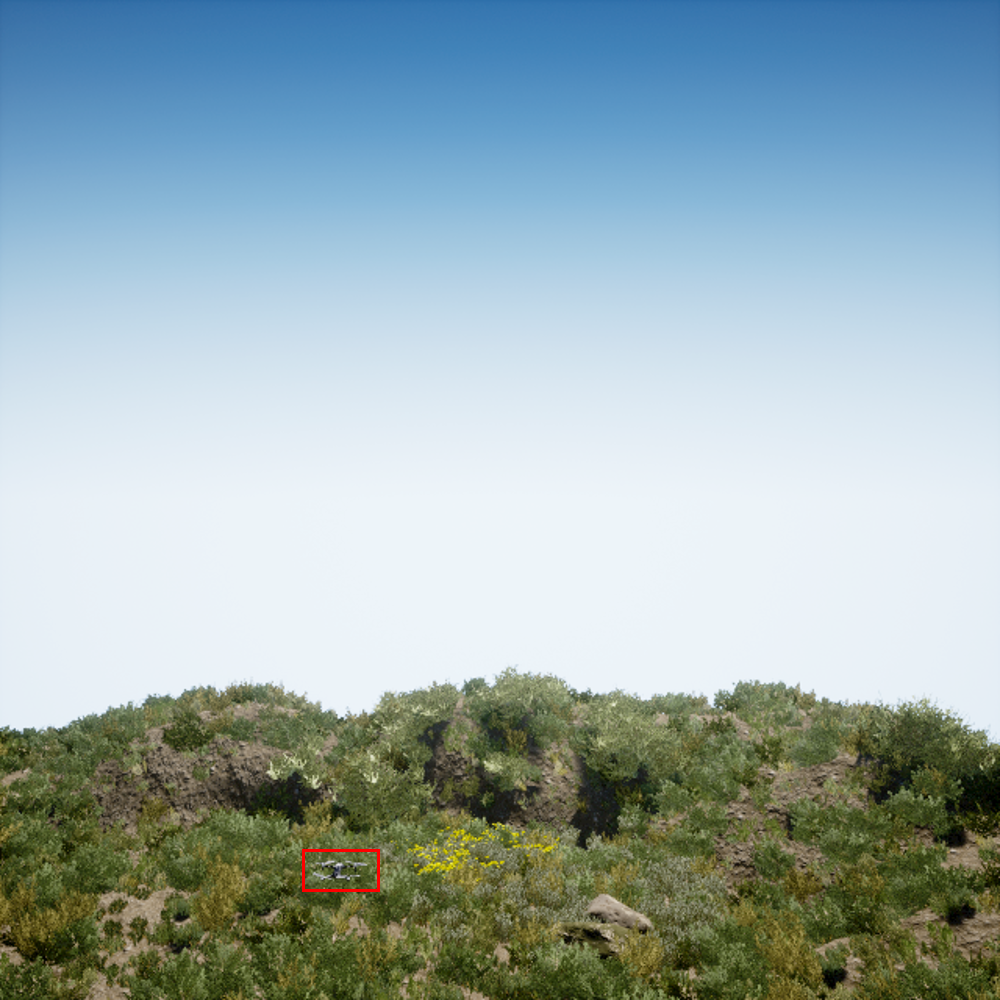}
		}
	
		\subfigure[Mountain Lake]{
			\centering
			\includegraphics[width=0.35\columnwidth, height=0.3\columnwidth]{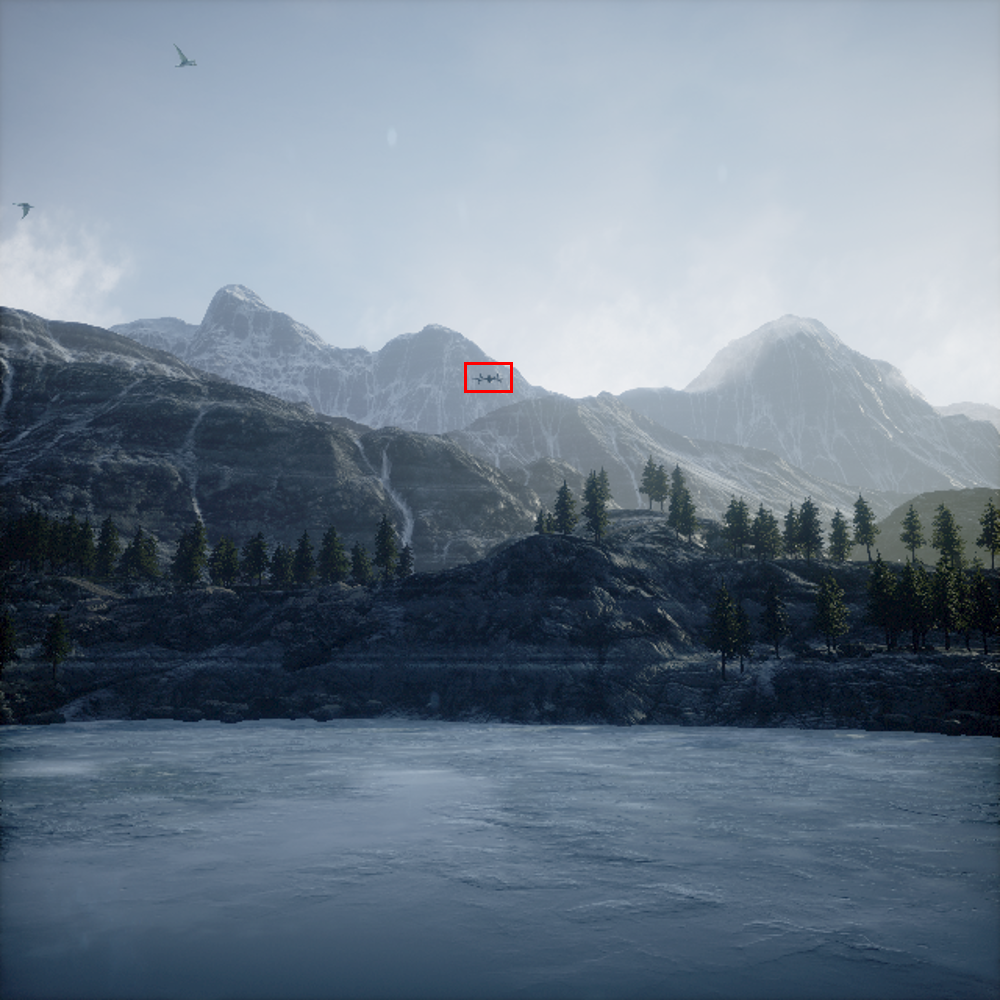}
		}
		\subfigure[Palace]{
			\centering
			\includegraphics[width=0.35\columnwidth, height=0.3\columnwidth]{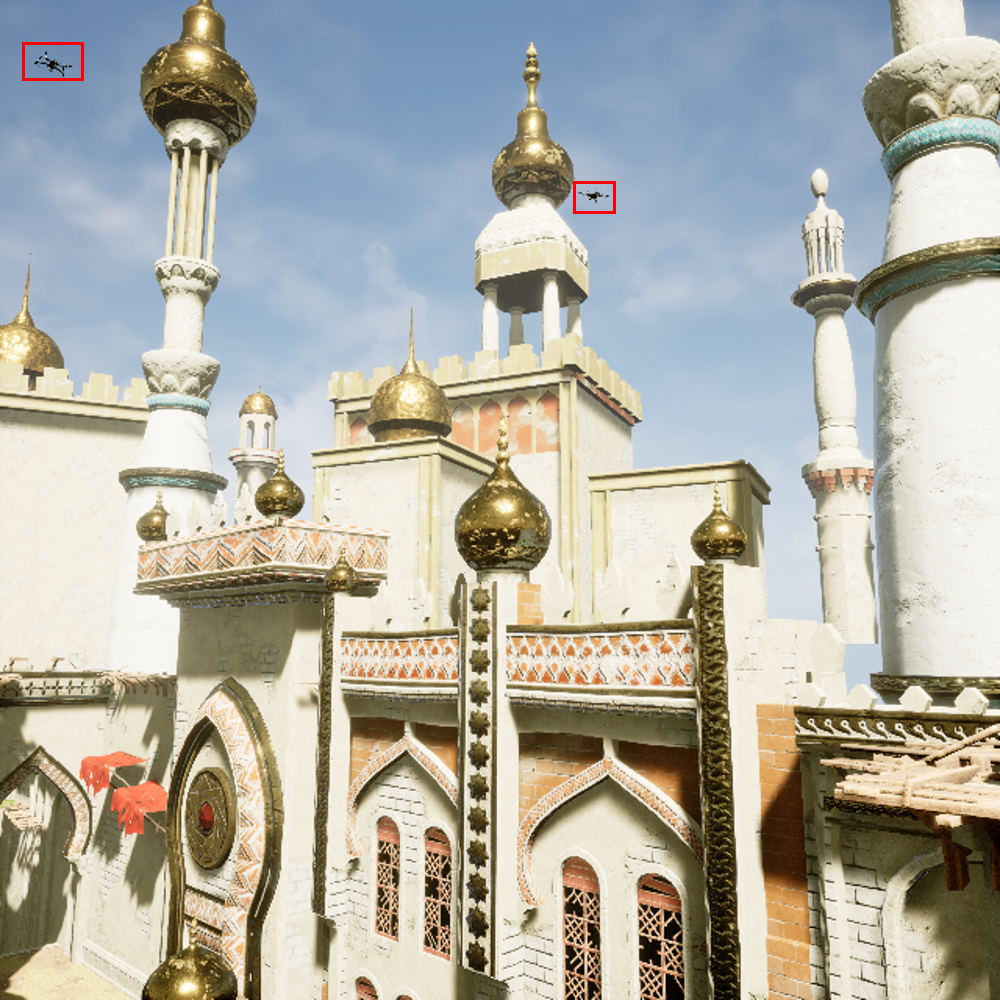}
		}
		\subfigure[Seaside Temple]{
			\centering
			\includegraphics[width=0.35\columnwidth, height=0.3\columnwidth]{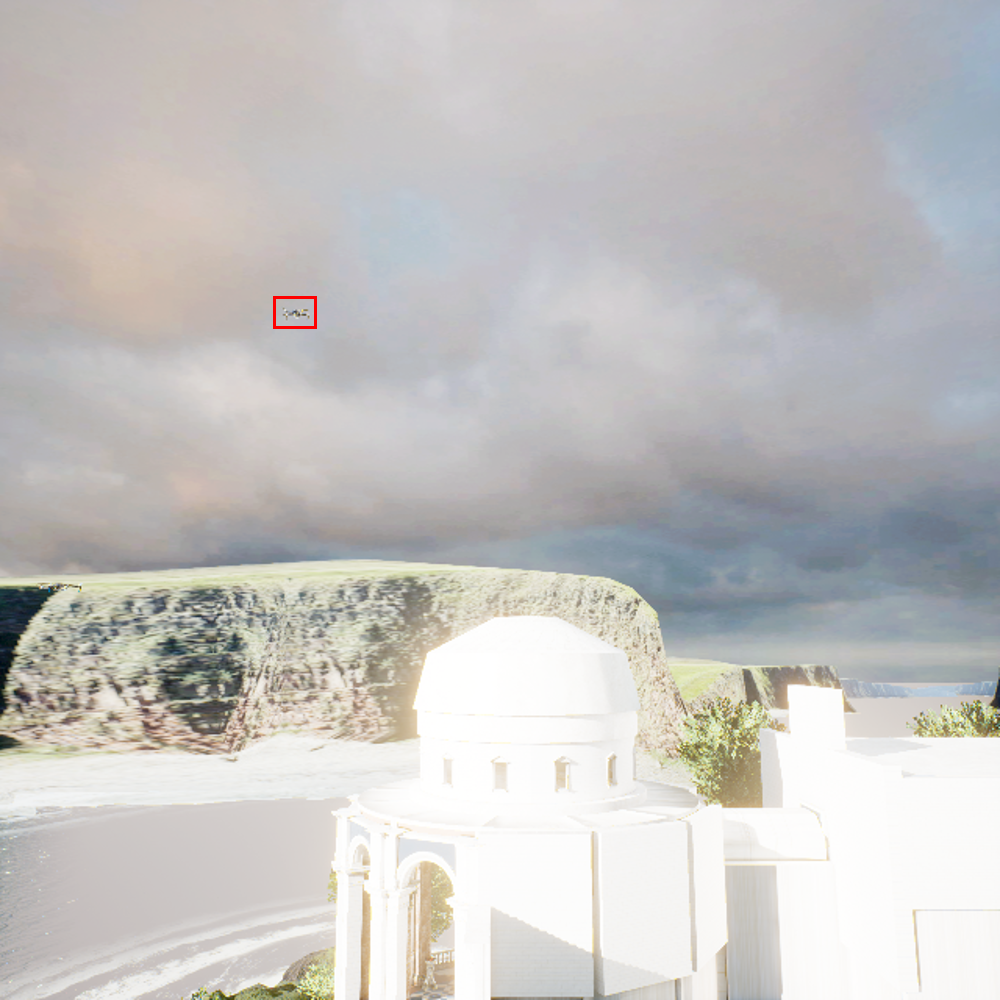}
		}
		\subfigure[Winter Town]{
			\centering
			\includegraphics[width=0.35\columnwidth, height=0.3\columnwidth]{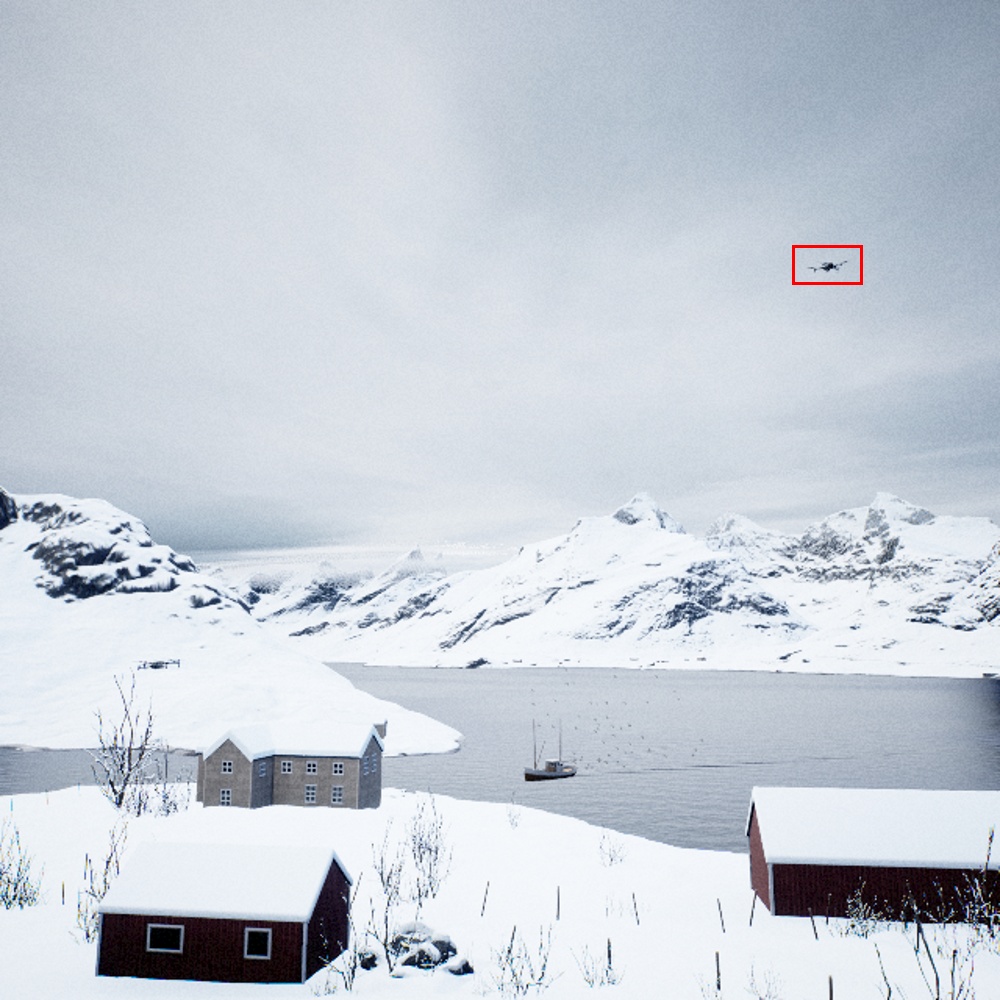}
		}
		
		\subfigure[Parrot A.R. Drone 2.0]{
			\centering
			\includegraphics[width=0.35\columnwidth]{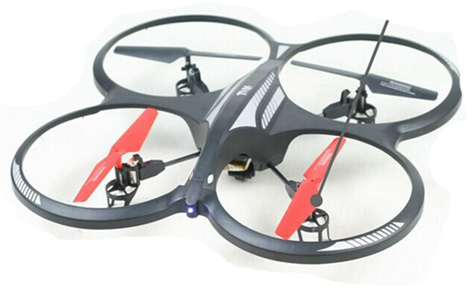}
		}
		\subfigure[DJI Inspire I]{
			\centering
			\includegraphics[width=0.35\columnwidth]{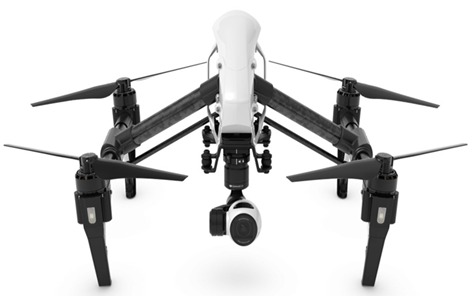}
		}
		\subfigure[DJI Mavic 2 Pro]{
			\centering
			\includegraphics[width=0.35\columnwidth]{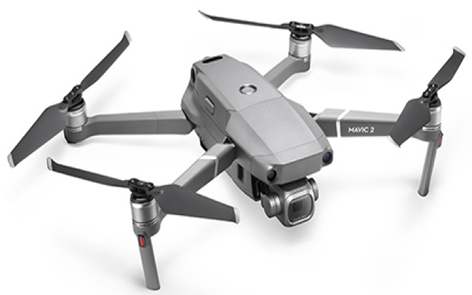}
		}
		\subfigure[DJI Phantom 4 Pro]{
			\centering
			\includegraphics[width=0.35\columnwidth]{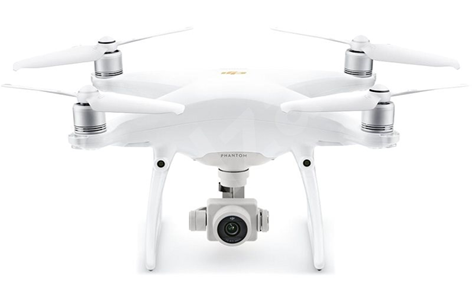}
		}
		\caption{(a)$ \sim $(h): Example images of SimUAV in 8 scenes, each has been annotated by red boxes. Note that the multi-rotor model in each image is DJI Mavic 2 Pro. (i)$ \sim $(l): All 3D models for multi-rotors we used in simulation.}
		\label{examples_SD}
	\end{figure*}

	SimUAV is a simulation dataset for UAV small object detection. It contains 29,568 images in 8 scenes, and each scene covers 4 multi-rotor models. Figure \ref{sim_prop} shows the proportion of each scene and model, indicating that each model accounts for the same proportion in the same scene, and there are four scenes(street, trees, mountain lake, winter town) that occupy more percentages because of complexity and large size of the environments. Figure \ref{examples_SD} displays the examples and 3D models of multi-rotors.
	
	We apply Airsim and UE4 as our simulation platform. Airsim \footnote{https://microsoft.github.io/AirSim/} is an open-source simulator for drones, built on Epic Games’ Unreal Engine 4\footnote{https://www.unrealengine.com/} as a platform for AI research. Airsim allows users to fly multi-rotors in UE4 environments at various velocities and on expected paths. Besides, Airsim allow to substitute any 3D multi-rotor model for the default model, and UE4 provides various environments or scenes to download in Epic Store.
	
\begin{figure}[htbp]
		\centering
		\includegraphics[width=1\columnwidth]{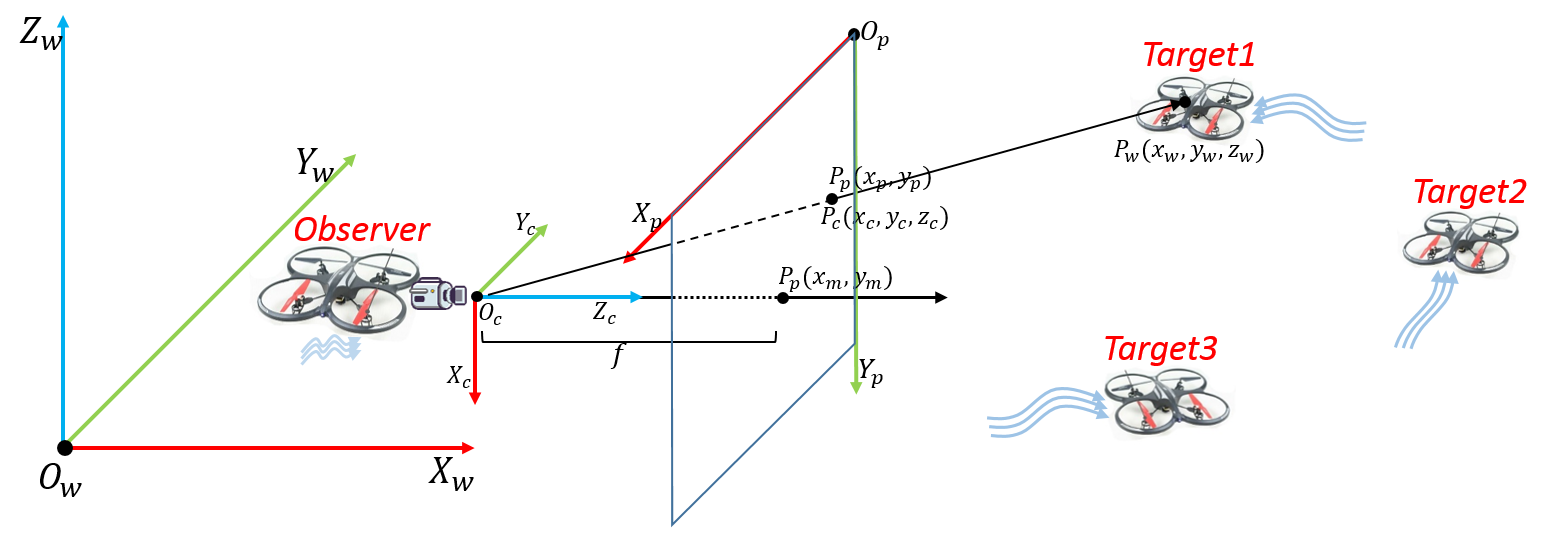}
		\caption{Simulation method: the observer shoots the target multi-rotors with onboard cameras}
		\label{simulation method}
	\end{figure}

	The generation of the SimUAV dataset is implemented by Airsim Python API. As described in Figure \ref{simulation method}, we create four multi-rotors, one as the observer, which carries several cameras(we only depict one camera) and others as the targets flying at a certain distance from the observer. Targets are expected to be captured by one or more cameras, and consequently, the camera images will be stored. Note that we keep the observer vibrating properly during simulation in order to imitate recording by human hand.
	
	Simultaneously the labels of each image can be computed, since the positions of targets and attitudes of cameras are available by Airsim during recording. Taking Target1 in Figure \ref{simulation method} as an example, there are three frames involved: world frame, camera frame, and pixel frame, respectively represented by $O_wX_wY_wZ_w$, $O_cX_cY_cZ_c$, and $O_pX_pY_p$. Therefore, we express the position of Target1 in three frames as $P_w(x_w,y_w,z_w)$, $P_c(x_c,y_c,z_c)$, and $P_p(x_p, y_p)$ respectively. Obviously, $P_p(x_p, y_p)$ is what we need for annotation and can be computed by the following formulas \ref{formula1} and \ref{formula2}:
	\begin{equation}
		\begin{bmatrix}
			x_c\\
			y_c\\
			z_c
		\end{bmatrix}=R\begin{bmatrix}
			x_w\\
			y_w\\
			z_w
		\end{bmatrix} + t
		\label{formula1}
	\end{equation}
	where $R\in \{R|R^TR=I_3, det(R)=1\}$ is the rotation matrix with respect to the rotation from $O_wX_wY_wZ_w$ to $O_cX_cY_cZ_c$ and $t$ is the translation vector equal to the camera position in the world frame(both $R$ and $t$ can be obtained by Airsim with a little computation);
	\begin{equation}
		\begin{cases}
			x_p = y_m - f\frac{y_c}{z_c}\\
			y_p = x_m + f\frac{x_c}{z_c}
		\end{cases}
		\label{formula2}
	\end{equation}
	where $f$ is the focal length, and$P_p(x_m, y_m)$ is the center position in pixel frame, for example $P_m(320, 320)$ in a 640*640 image.
	
	Overall, we generate all the annotated images in various scenes and drone models, forming the SimUAV dataset.
	
	\subsection{Transfer Learning}
	\label{method_tl}
	Through transfer learning, neural networks learn features from simulation data first and then these features can be applied to real-world data to reduce the demand for huge human annotations. With the aid of simulation data and real-word data, we need to reduce the error gap (also named generalization error) between source domain and target domain to be as small as possible. The source domain error function $\epsilon_{\widehat{P}}(f)$ adopted 01-loss is given by:
	\begin{equation}
		\epsilon_{\widehat{P}}(f)=\frac{1}{n} \sum_{i=1}^{n}\left[f\left(\mathbf{x}_{i}\right) \neq y_{i}\right]=\mathbb{E}_{(\mathbf{x}, y) \sim P}[f(\mathbf{x}) \neq y]
	\end{equation}
	where $\widehat{P}=\left\{\left(\mathbf{x}_{i}, y_{i}\right)\right\}_{i=1}^{n}$ represents the source domain's distribution, $f$ is the non-linear map function. Target error $\epsilon_{\widehat{Q}}(f)$ is given by:
	\begin{equation}
		\epsilon_{\widehat{Q}}(f)=\mathbb{E}_{(x', y') \sim Q}[f(\mathbf{x'}) \neq y']
	\end{equation}
	where $(x',y')\sim \widehat{Q}$ denotes data sampled from the target domain $Q$. Therefore, generalization error is given by:
	\begin{equation}
		\begin{aligned}
			\epsilon_{G}(f)&=\epsilon_{\widehat{Q}}(f)-\epsilon_{\widehat{P}}(f)\\
			&=\mathbb{E}_{(x', y') \sim Q}[f(\mathbf{x'}) \neq y']-\mathbb{E}_{(\mathbf{x}, y) \sim \hat{P}}[f(\mathbf{x}) \neq y]
		\end{aligned}
	\end{equation}
	
	In other words, the goal of transfer learning is to utilize the knowledge of the source domain in the case of $P\neq Q$ to improve the prediction performance of the non-linear map function $f(\cdot)$ of target task. In \cite{yosinski2014transferable}, the experiment proved that common features could be obtained through first three layers of the network, and transfer learning improved the model's performance with fine-tuning of the original network. Moreover, this experiment also indicated that the performance of the deep transfer network is better than the network with random initialization. Besides, number of layers for transfer learning would have a significant influence on the speed of learning and optimization process.

	Transfer learning adopted in this framework includes two stages as above-mentioned: pre-training stage and fine-tuning stage. The amount of target data and the similarity between source data and target data determine the way of transfer learning. Once the amount of target data is limited, low-level feature extraction part of pre-trained model should be retained and only weights of high-level layers could be modified. Once the amount of target data is relatively large or there is a distinctively difference between source data and target data, it is necessary to change the structure and retrain whole weights of pre-trained model. To reveal the effect of learning pattern and quantity of dataset, different experiments have been conducted and related results can be found in section \ref{experiment}.
	
	\subsection{Adaptive Fusion}
	\label{method_af}
\begin{figure*}[t]
		\centering
		\includegraphics[width=0.8\linewidth]{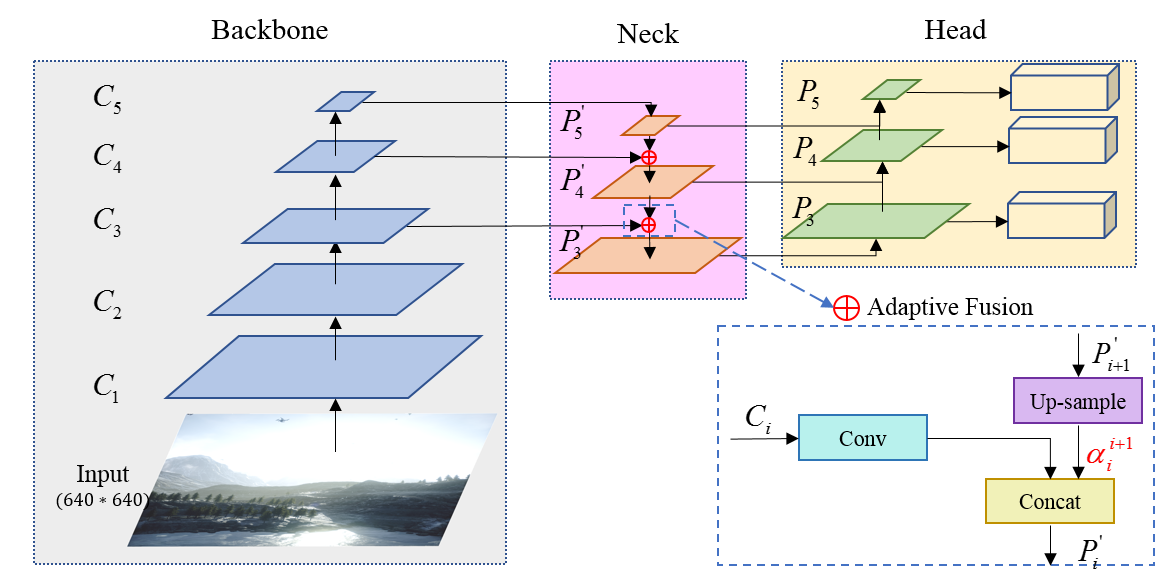}
		\caption{The network structure of AF-YOLO model.}
		\label{AF-YOLO}
	\end{figure*}

	The YOLO v5 model illustrated in Figure \ref{AF-YOLO} contains three parts: Backbone, Neck and Head, detailed information can be found in github \footnote{https://github.com/ultralytics/yolov5}. PANet has been integrated into the neck part of YOLO v5 model to realize feature fusion of different hierarchies. {$P_5'$, $P_4'$, $P_3'$} denote different feature levels generated by top-down path, from $P_5'$ to $P_3'$, the spatial size is up-sampled with factor 2 each time. {$P_3$, $P_4$, $P_5$} represent feature maps generated by path aggregation corresponding to {$P_3'$, $P_4'$, $P_5'$}.
	As the way to fuse features of different layers determine the performance of object detection, we propose adaptive fusion mechanism to control the degree of fusion between same-sized features such as upsampling of $P_5'$ and $C_4$. As shown in the right bottom part of Figure \ref{AF-YOLO}, adaptive fusion mechanism merges top-down convolution feature of $C_i$ and upsampling feature of $P_{i+1}'$ together, and $P_i'$ is given by:
	\begin{equation}
		P_i'=\alpha_i^{i+1}\times f_{up-sample}(P_{i+1}') \oplus f_{conv}(C_i)
	\end{equation} 
	where $\oplus{+}$ denotes concatenate operation. In fact, since features extracted by different feature maps are almost different, feature $P_i'$ learn to combine shallow feature $C_i$  and deep feature $P_{i+1}'$ together with the new adaptive fusion mechanism. The coefficients change from fixed to learnable make it possible to achieve distinguished performance on object detection of different scales.
	
	\section{Experiment}
	\label{experiment}
	In this section, we firstly describe the partitions of our datasets and the evaluation criteria used in experiments. Then several experiments are conducted to validate proposed methods, including the validation of simulated dataset we generated, the performance of transfer learning, and the improvement brought by AF-YOLO model. For each experiment in this paper, we employ YOLO v5 as our baseline detector implemented by Pytorch with training epoch and batch size set to 200 and 16 respectively. 
	
	\subsection{Dataset and Evaluation}
	\subsubsection{Dataset Partitions}
	We conduct experiments on both the real-world dataset and the simulated dataset. In order to explore influence of low-resource on performances, we divide both two datasets. According to the description of UAV small object detection, we firstly generate a test dataset which contains 3322 samples from the real-world dataset with only small targets(pixels $\leq32^2$). Then we split uniformly the rest of real-world dataset into three parts and take the first part as RD1, the first two parts as RD2, and all three parts as RD3. Similarly, we divide the whole simulated dataset into three parts and construct SD1, SD2, and SD3. Table \ref{data_partitions} shows all subsets used in experiments with corresponding target sizes.
	
	\begin{table}[htb]
		\centering
		\resizebox{\linewidth}{!}{
			\begin{tabular}{llllllll}
				\toprule   
				&Subset&$0\sim8^2$&$8^2\sim16^2$&$16^2\sim32^2$&$\geq32^2$&total\\
				\midrule
				\multirow{4}{*}{RD}&Test&405&2086&831&-&3322\\
				&RD1&133&965&1957&2490&5545\\
				&RD2&304&2240&4673&5719&12936\\
				&RD3&437&3205&6630&8209&18481\\
				\midrule
				\multirow{3}{*}{SD}&SD1&-&1369&8122&314&9805\\
				&SD2&-&2851&14164&582&17597\\
				&SD3&-&5992&22931&645&29568\\
				\bottomrule
		\end{tabular}}
		\caption{Number of images of each subset over different scales. Noting real-world dataset holds a higher percentage of large objects than simulated dataset.}
		\label{data_partitions}
	\end{table}
	
	\subsubsection{Evaluation Criteria}
	Since there is only one category in our task, we use the commonest metric \emph{Average Precision}(AP) \cite{Mark_pascal_challenge} to evaluate the performance. A prediction $(b, conf)$ returned by a detector, where $b$ is the bounding box and $conf$ the confidence, is regarded as a True Postive(TP) if the overlap ratio IOU(Intersection Over Union) \cite{Mark_pascal_challenge} between the predicted $b$ and the ground truth one is not smaller than 0.5. Based on the TP detection, the different precision and recall pairs $(P(\beta), R(\beta))$ can be computed by various confidence threshold $\beta$. Hence the precision at a certain recall can be obtained based on different $(P, R)$ pairs \cite{Olga_imageNet_challenge}. Finally, the AP can be derived from the following formulas \cite{Mark_pascal_challenge}:
	\begin{equation}
		AP=\frac{1}{11}\underset{r\in\{0,0.1,...,1\}}{\sum}p_{interp}(r)
	\end{equation}
	where $p_{interp}(r)$ is the interpolated precision at each recall level $r$:
	\begin{equation}
		p_{interp}(r)=\underset{\tilde{r}:\tilde{r}\geq r}{\max}p(\tilde{r})
	\end{equation}
	and $p(\tilde{r})$ is the precision at a certain recall $\tilde{r}$.
	
	\subsection{Validation of SimUAV}
	\label{exp_dataset}
	To validate whether simulated dataset is qualified as the auxiliary training resource for our task, we design an experiment that YOLO v5 model is trained as the baseline on all RDs and SDs. Table \ref{tb_simuav_valid} shows the evaluation results on test dataset of all training sets in the view of different target sizes. It can be found that all APs of SDs and RDs lie in the range from 0.5 to 0.6 and the gap narrows when trained on SD3. Considering the different distributions of RDs and SDs, we think simulated dataset is adequate to act as the auxiliary for real-world dataset. Note that SD2 get the best performance for target size of $16^2\sim32^2$ while table \ref{data_partitions} suggests that targets within $16^2\sim32^2$ take a dominant proportion in SDs. Obviously, we need a method to make full use of simulated dataset.
	\begin{table}[htb]
		\centering
		\resizebox{\linewidth}{!}{
			\begin{tabular}{lllll}
				\toprule
				Train Set&$AP_{0\sim8^2}$&$AP_{8^2\sim16^2}$&$AP_{16^2\sim32^2}$&$AP_{0\sim32^2}$\\ 
				\midrule
				RD1&0.039&0.589&0.751&0.565\\
				RD2&0.255&0.602&0.765&0.596\\
				RD3&0.195&0.600&0.758&0.592\\
				SD1&0.163&0.574&0.596&0.519\\
				SD2&0.005&0.496&\textbf{0.769}&0.506\\
				SD3&0.109&0.570&0.724&\textbf{0.559}\\
				\bottomrule
		\end{tabular}}
		\caption{Performances of baseline model(YOLO v5) of different training datasets.}
		\label{tb_simuav_valid}
	\end{table}
	
	\subsection{Transfer Learning}
	\label{exp_tl}
	We introduce transfer learning to our task with two main goals:
	\begin{itemize}
		\item improve the performance on real-world dataset by transfering general knowledge for UAV small object detection from simulated dataset to real-world dataset.
		\item improve data efficiency in case of insufficient real-world data.
	\end{itemize}
	
\begin{figure*}[t]
		\centering
		\subfigure[]{
			\centering
			\includegraphics[width=0.6\columnwidth]{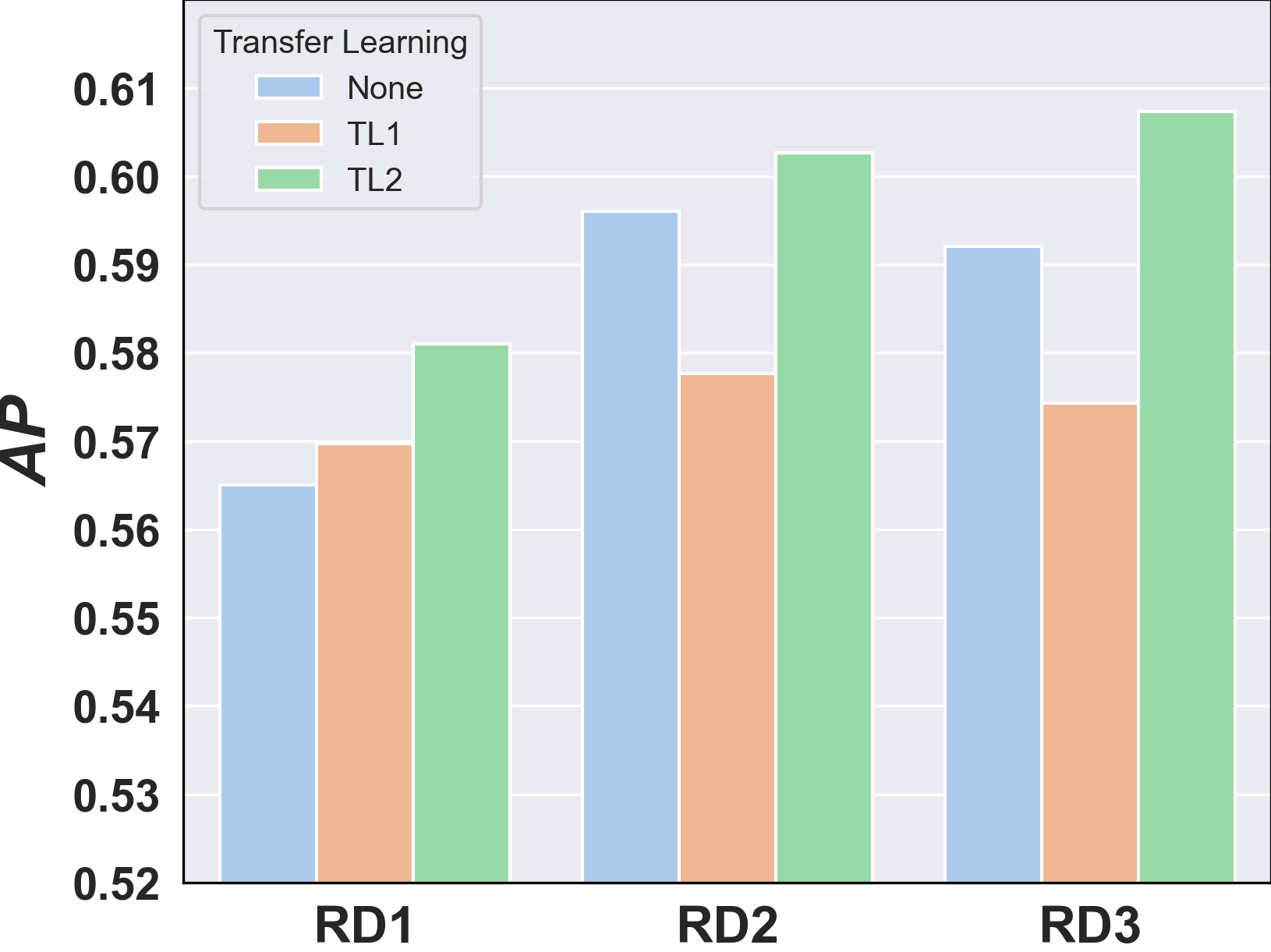}
			\label{fg_exp_tl_a}
		}
		\subfigure[]{
			\centering
			\includegraphics[width=0.6\columnwidth]{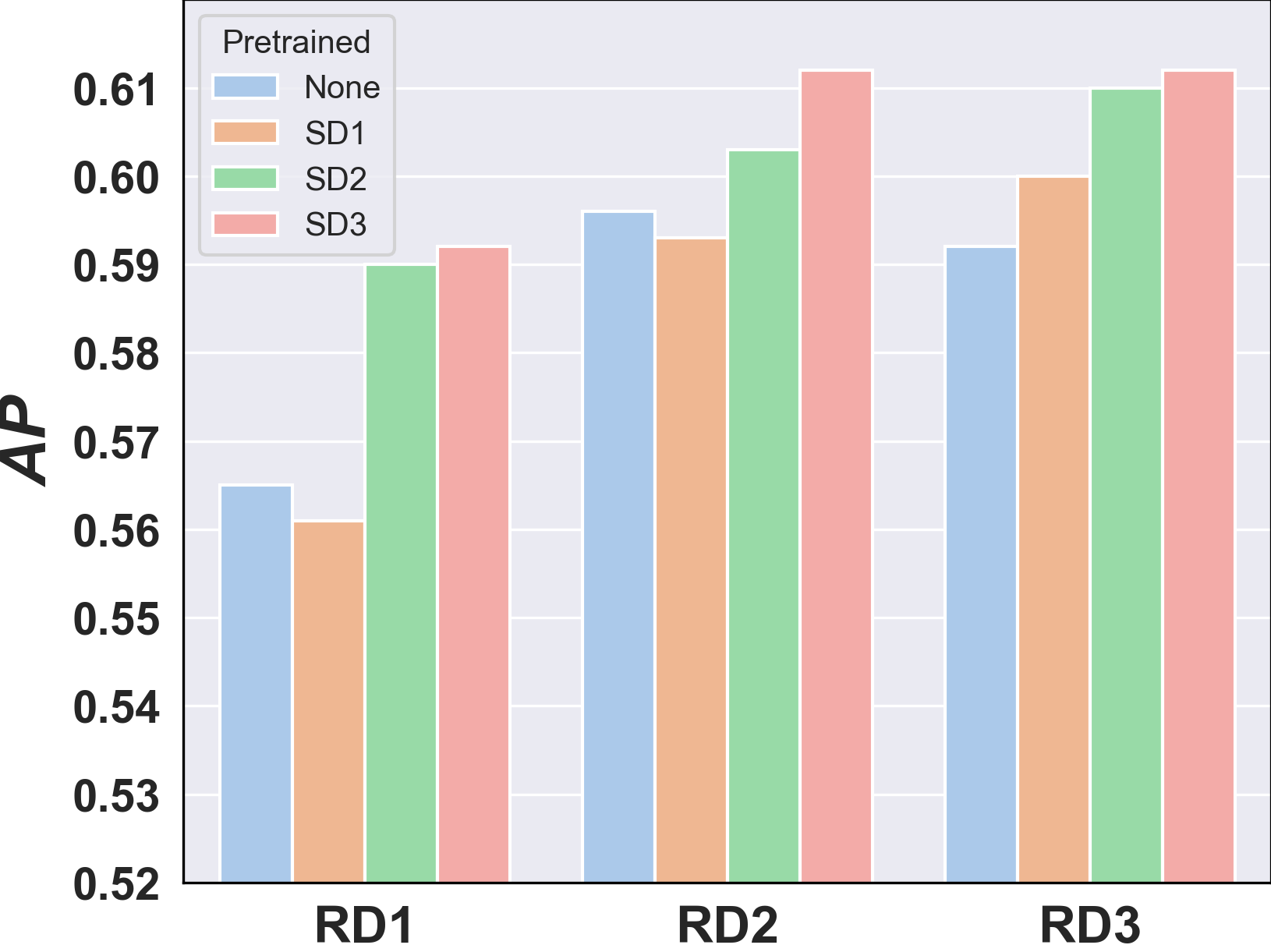}
			\label{fg_exp_tl_b}
		}
		\subfigure[]{
			\centering
			\includegraphics[width=0.6\columnwidth]{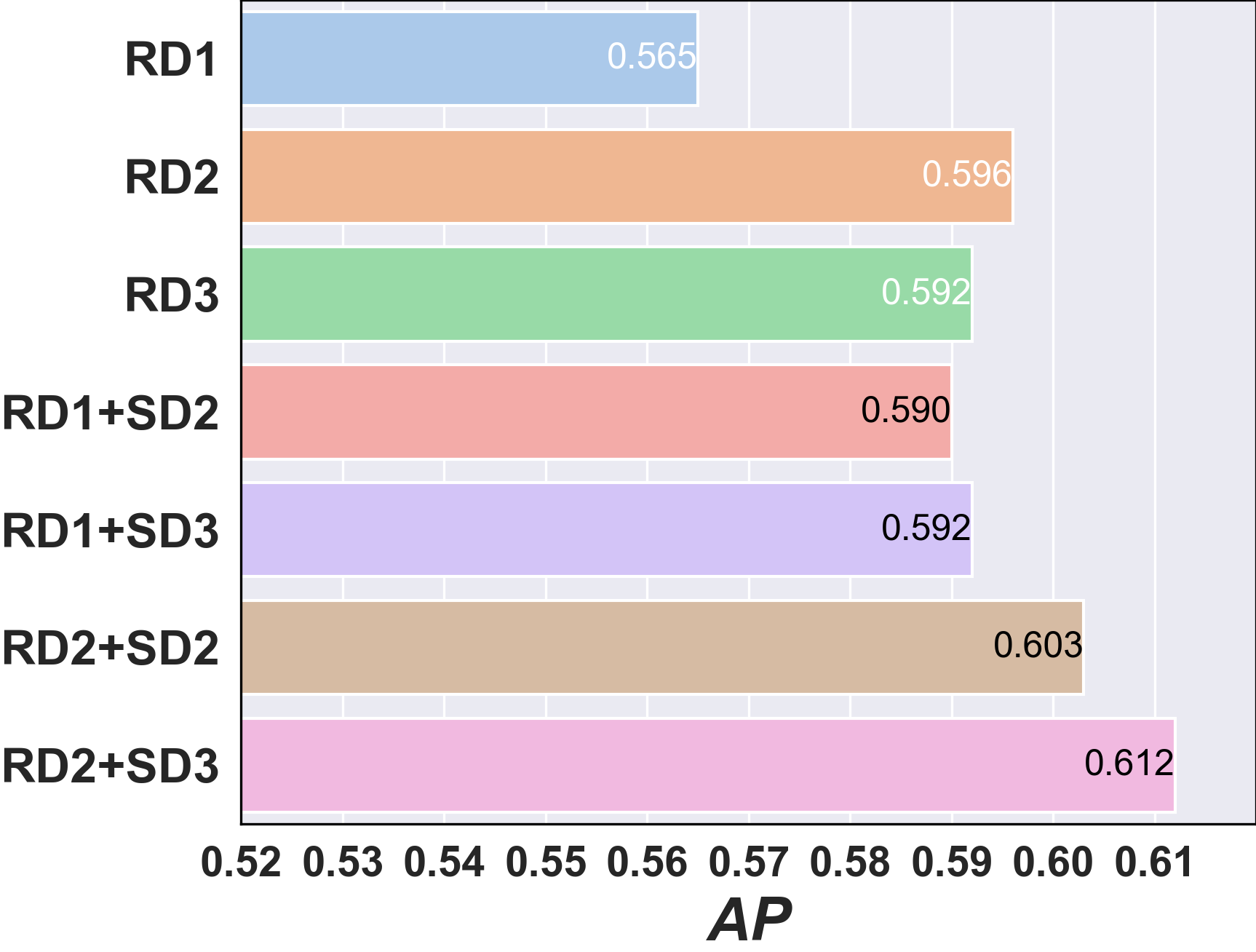}
			\label{fg_exp_tl_c}
		}
		\caption{Diagrams derived from Tabel \ref{tbl_exp_tl}: (a) overall APs of two methods of transfer learning on different RDs, showing the improvement brought by TL2 but the failure of TL1; (b) APs of TL2 pretrained on different SDs and finetuned on different RDs, implying that SDi with enough volume is helpfull; (c) APs of some combinations of RDs and SDs, indicating that the performance of low-resource real-world data equiped with enough simulated data gets close to performances of rich real-world data(e.g. RD1+SD3 vs. RD3) or even surpasses(e.g. RD2+SD2 vs. RD3)}
		\label{fg_exp_tl}
	\end{figure*}

	As mentioned in subsection \ref{method_tl}, we apply two methods of transfer learning denoted by TL1 and TL2 in oder to seek a saving training for finetuning. We freeze the backnone part of the baseline detector when finetuning for TL1, while we retrain all parts for TL2. Both TL1 and TL2 are pretrained on simulated dataset and finetuned on real-world dataset. There are 18 combinations for experiment setting, considering 2 methods of transfer learning, 3 SDs and 3 RDs. Table \ref{tbl_exp_tl} shows all results of experiments about transfer learning and gives detail performances of different target sizes as usual. For convenience, we produce three digrams in Figure \ref{fg_exp_tl} which can be derived from Tabel \ref{tbl_exp_tl}.
	
	\begin{table}[htb]
		\resizebox{\linewidth}{!}{
			\centering
			\begin{tabular}{lllll}
				\toprule
				&$AP_{0\sim8^2}$&$AP_{8^2\sim16^2}$&$AP_{16^2\sim32^2}$&$AP_{0\sim32^2}$\\ 
				\midrule
				(RD1)&0.039&0.589&0.751&0.565\\
				RD1+TL1+SD1&0.125&0.587&0.748&0.570\\
				RD1+TL1+SD2&0.120&0.578&0.766&0.577\\
				RD1+TL1+SD3&0.078&0.572&0.752&0.562\\
				RD1+TL2+SD1&0.041&0.586&0.768&0.561\\
				RD1+TL2+SD2&0.106&0.604&0.792&0.590\\
				RD1+TL2+SD3&0.103&0.603&0.775&0.592\\
				\hline
				(RD2)&0.255&0.602&0.765&0.596\\
				RD2+TL1+SD1&0.102&0.591&0.763&0.579\\
				RD2+TL1+SD2&0.144&0.593&0.775&0.590\\
				RD2+TL1+SD3&0.116&0.565&0.764&0.564\\			
				RD2+TL2+SD1&0.138&0.599&0.789&0.593\\
				RD2+TL2+SD2&0.151&0.614&0.793&0.603\\
				RD2+TL2+SD3&0.182&0.619&0.797&0.612\\
				\hline
				(RD3)&0.195&0.600&0.758&0.592\\
				RD3+TL1+SD1&0.148&0.596&0.760&0.584\\
				RD3+TL1+SD2&0.114&0.591&0.770&0.586\\
				RD3+TL1+SD3&0.090&0.562&0.761&0.553\\
				RD3+TL2+SD1&0.141&0.608&0.793&0.600\\
				RD3+TL2+SD2&0.156&0.619&0.808&0.610\\
				RD3+TL2+SD3&0.238&0.620&0.789&0.612\\
				\bottomrule
		\end{tabular}}
		\caption{The performance of models with the transfer learning method. SDi: pre-training on SDi; RDi: fine-tuning on RDi; TL1: The first method of transfer learning, which freeze the backbone of baseline model during fine-tuning; TL2: The second method of transfer learning, which trains all parameters during fine-tuning.}
		\label{tbl_exp_tl}
	\end{table}
	
	Digram \ref{fg_exp_tl_a} shows overall APs of two methods of transfer learning finetuned on different RDs (averaging over SDs). We find that TL1 has only a slight increase as finetuned on RD1 compared with training from scrach and works worse on RD2 and RD3. Whereas TL2 brings obvious improvements finetuned on all RDs. We attribute these results to the different distributions of real-world dataset and simulated dataset. For example there are four UAV models in simulated dataset but only one in real-world dataset. Besides, simulated dataset has a uniform image size of 1920*1920 while most images in real-world is 1920*1080, which can produce difference when reshaped to 640*640 as the inputs of detectors. Therefore, we choose TL2 as the default transfer learning method for follow-up analyses in the rest of this section.
	
	Digram \ref{fg_exp_tl_b} presents performances pretrained on different SDs and finetuned on different RDs. There is an ascending of performances as the volume of SDi goes up, indicating that general knowledge for UAV small object detection could be extracted with enough simulated dataset and performance can be improved on real-world dataset. So far transfer learning help us achieve the first goal mentioned above.

	Digram \ref{fg_exp_tl_c} interprets that the low-resource of real-world dataset can be resolved by transfer learning with sufficient simulated dataset. We display APs of some combinations of RDs and SDs and find that with less real-world data and enough simulated data the performance gets close to ones of more real-world dataset(e.g. RD1+SD3 vs. RD3) or even exceeds(e.g. RD2+SD2 vs. RD3). The second goal has been reached as well.
	
	\subsection{Adaptive Fusion}
	\label{exp_af}
	Aiming to obtain a better fusion prone for small object, we propose the AF-YOLO model which introduce adaptive fusion mechanism in top-down path of the neck block described in subsection \ref{method_af}. An experiment is conducted to study the effect of adaptive fusion. We train our AF-YOLO model on real-world dataset(RD1) and make a comparison with baseline models of some different fixed fusion coefficients.
	
	\begin{table}[htbp]
		\centering
		\resizebox{\linewidth}{!}{
			\begin{tabular}{lllll}
				\toprule
				&$AP_{0\sim8^2}$&$AP_{8^2\sim16^2}$&$AP_{16^2\sim32^2}$&$AP_{0\sim32^2}$\\ 
				\midrule
				$\alpha_4^5=\alpha_3^4=0.00$&0.266&0.608&0.804&0.620\\
				$\alpha_4^5=\alpha_3^4=0.25$&0.178&0.609&0.784&0.605\\
				$\alpha_4^5=\alpha_3^4=0.50$&0.168&0.602&0.769&0.596\\
				$\alpha_4^5=\alpha_3^4=0.75$&0.0879&0.571&0.777&0.565\\
				$\alpha_4^5=\alpha_3^4=1.00$&0.0394&0.589&0.751&0.565\\
				adaptive $\alpha_4^5$ and $\alpha_3^4$(AF-YOLO) &$\bm{0.346}$&$\bm{0.615}$&$\bm{0.811}$&$\bm{0.627}$\\
				\bottomrule
		\end{tabular}}
		\caption{Performances of different fusion coefficients. Each model is trained on RD1. $\alpha_4^5=\alpha_3^4=1.00$ as the default setting in the baseline model means a combination of full deep features and shallow features, while $\alpha_4^5=\alpha_3^4=0.00$ means no influence of deep features. Adaptive $\alpha_4^5$ and $\alpha_3^4$ act as learnable parameters in AF-YOLO and finally converge to -0.231 and 0.013 respectively in this experiment.}
		\label{tbl_exp_af}
	\end{table}

	Table \ref{tbl_exp_af} shows the result that fusion coefficients in the top-down path has a significant impact on small object detection. $\alpha_4^5=\alpha_3^4=1.00$ means full deep features are propagated in the top-down path and get the worst performance. Instead, $\alpha_4^5=\alpha_3^4=0.00$ means none of deep features involed but works best among all fixed fusion coefficients. As the fusion coefficients descend, there is an uptrend of performance for each target size in table \ref{tbl_exp_af}. More importantly, AF-YOLO model achieves the best performances over all target sizes and makes the overall AP increase up to 6.2\% compared with the default fusion coefficients. This experiment has proved that deep features do hinder shallow features for the task of small object detection and adaptive fusion coefficients learned by training produce the best fusion. 
	
\begin{figure*}[t]
		\centering
		\subfigure[An example containing a small-size UAV from the real-world dataset.]{
			\centering
			\includegraphics[width=0.97\columnwidth, height=0.5\columnwidth]{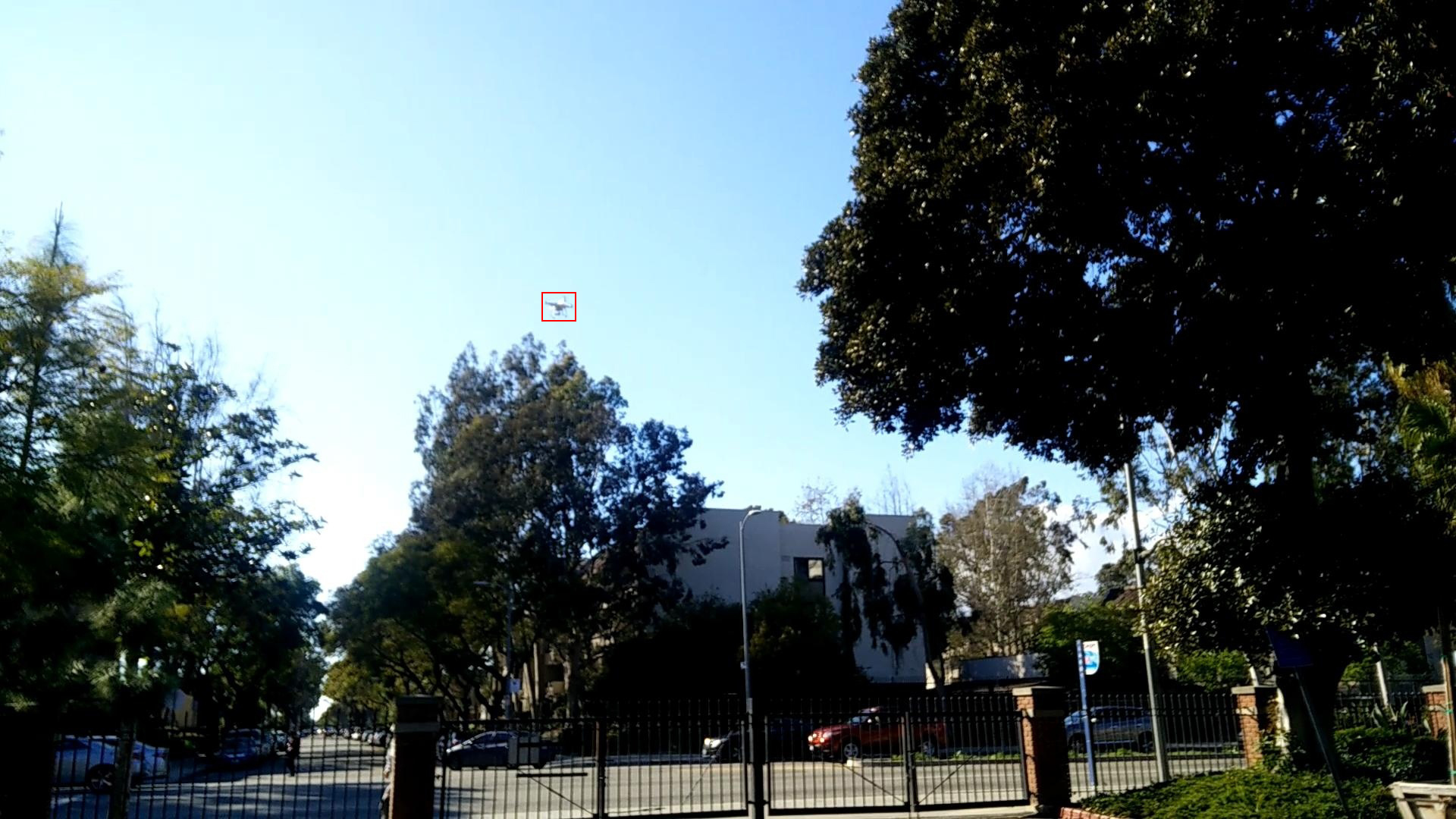}
			\label{a}
		}
		\subfigure[An example containing a small-size UAV from the simUAV dataset.]{
			\centering
			\includegraphics[width=0.97\columnwidth, height=0.5\columnwidth`]{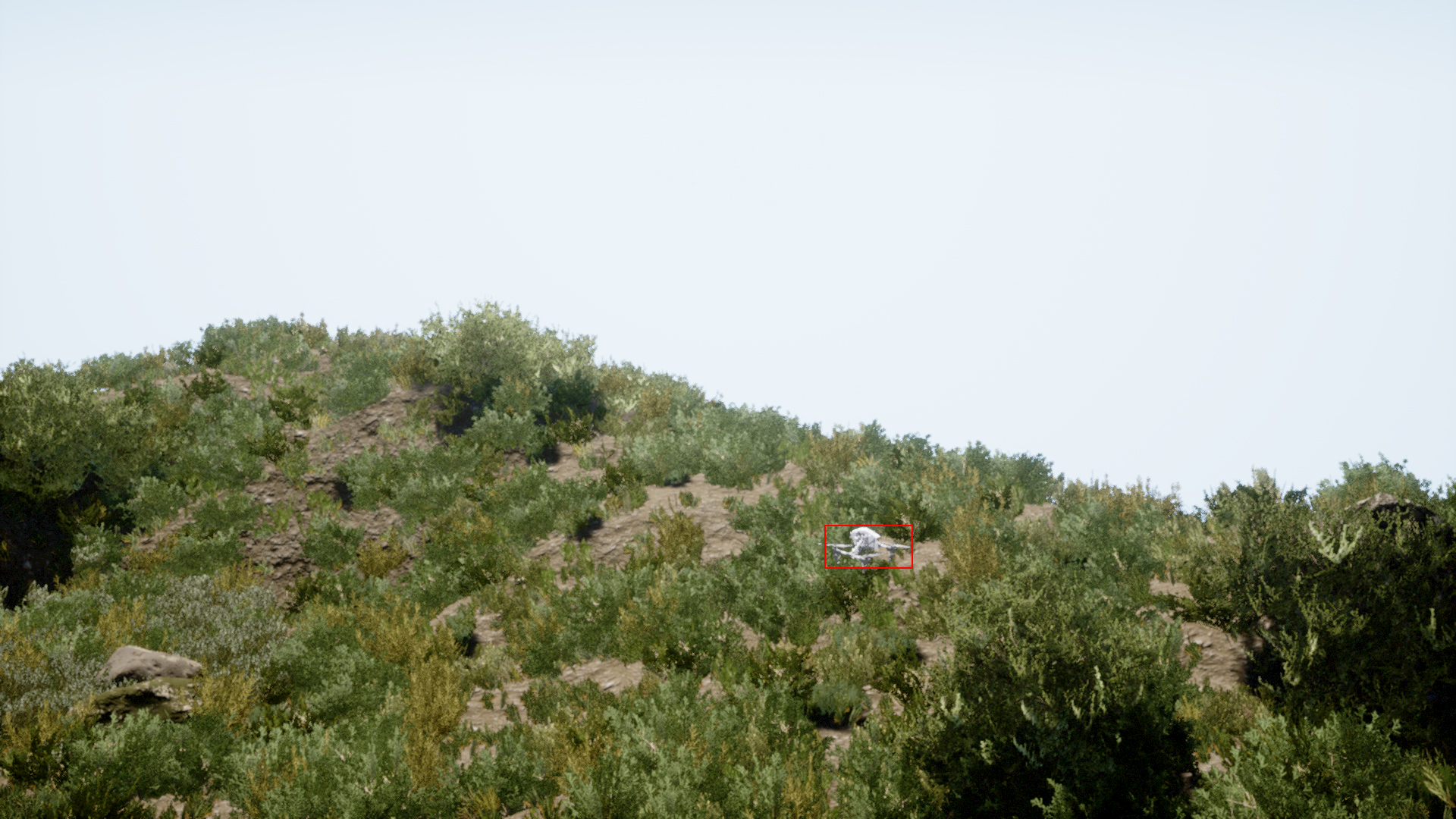}
			\label{b}
		}
		\subfigure[16 feature maps of (a) after passing through the original network or the improved network with the fusion coefficients. Row 1 presents 16 feature maps after passing through the original network. Row 2 presents 16 feature maps after passing through the improved network with the fusion coefficients.]{
			\centering
			\includegraphics[width=2\columnwidth]{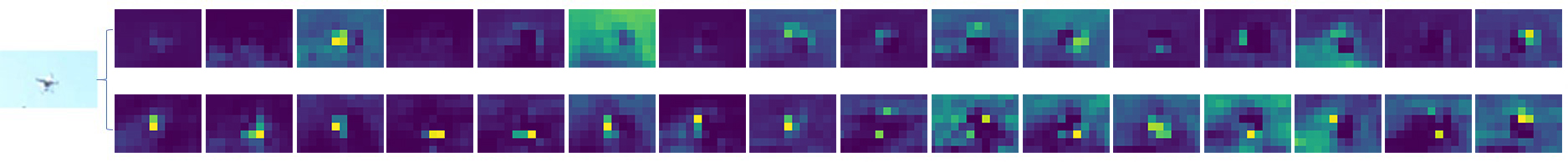}
		}
		
		\subfigure[16 feature maps of (b) after passing through the original network or the improved network with the fusion coefficients. Row 1 presents 16 feature maps after passing through the original network. Row 2 presents 16 feature maps after passing through the improved network with the fusion coefficients.]{
			\centering
			\includegraphics[width=2\columnwidth]{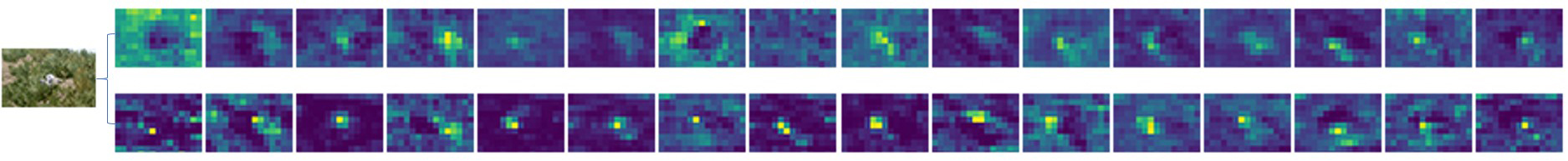}
		}
		\caption{The comparison of 16 feature maps from the same channels generated by original feature fusion and new feature fusion with fusion coefficients for small object detecting.}
		\label{comparision}
	\end{figure*}

	Figure \ref{comparision} show the visualization of feature maps containing small UAVs. It can be found that more distinctive feature maps could be obtained by AF-YOLO, which indicates that feature representation ability for small objects can be enhanced by adaptive fusion coefficients.
	
	\subsection{Transfer Leanrning with Adaptive Fusion}
\begin{figure*}[h]
		\centering
		\includegraphics[width=2\columnwidth]{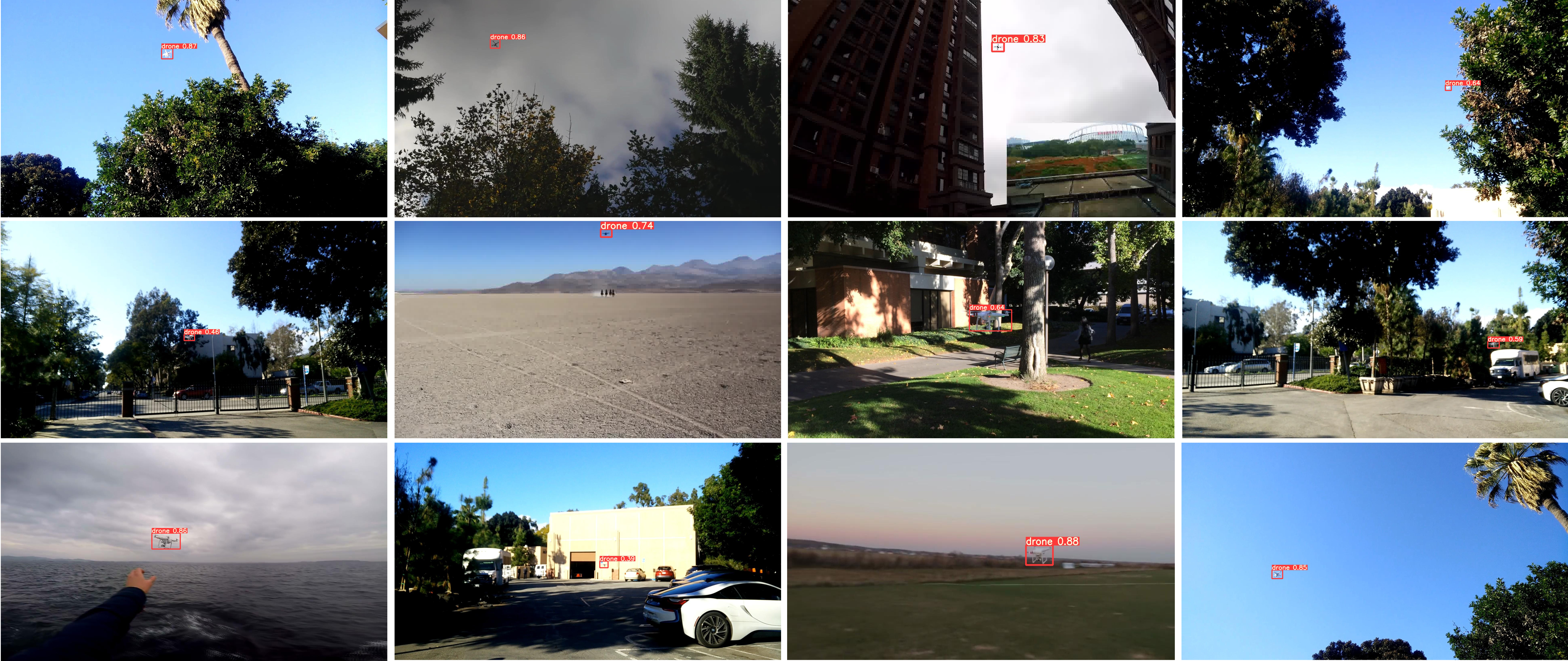}
		\centering
		\caption{Detection results with object detector combining transfer learning and the learnable coefficients, and all of them are on a small scale.}
		\label{Detection results}
	\end{figure*}

	Finally, we combine the methods of the second transfer learning way and adaptive fusion coefficients. Table \ref{the model combining transfer learning and the learnable coefficients} shows that the combination of transfer learning and adaptive fusion methods could improve the detection and recognition performance of UAV small objects effectively. Figure \ref{Detection results} shows examples of our detection results. Finally, an improvement of 7.1\% can be achieved by the comprehensive approach compared to the baseline model.
	\begin{table}[htbp]
		\centering
		\resizebox{\linewidth}{!}{
			\begin{tabular}{lllll}
				\toprule
				Real-world data&Simulation data&Transfer learning&$\bm{\alpha}$&$AP_{0\sim32^2}$\\ 
				\midrule
				RD1&-&-&-&0.565\\
				-&SD3&-&-&0.559\\
				RD1&SD3&\checkmark&-&0.592\\
				RD1&-&-&\checkmark&0.627\\
				RD1&SD3&\checkmark&\checkmark&0.636\\
				\bottomrule
		\end{tabular}}
		\caption{Performance of the model combining transfer learning and the adaptive fusion coefficients.}
		\label{the model combining transfer learning and the learnable coefficients}
	\end{table}

	\section{Conclusion}
	\label{conclusion}
	Aiming at insufficient of real UAV images, AirSim plugin is proposed to generate simulation dataset. With transfer learning, knowledge extracted from simulation data can be utilized to decrease the demand of real data and improve detection performance. Moreover, adaptive fusion mechanism proposed learns the appropriate way to fuse deep features and shallow features which results in a significant performance improvement. Our work presents a comprehensive approach for UAV small object detection. In the future, we will focus on expanding the dataset to cover more common UAV models, sensor types and real flight scenes. Furthermore, Cross-modal learning which invole different sensors(such as RGB, IR, Radar) should be explored to enchance the applicability of object deteciton algorithms over various weather conditions.
	
	\bibliographystyle{elsarticle-num}
	\bibliography{paper}
	
\end{document}